\documentclass[10pt,twocolumn,letterpaper]{article}

\usepackage[pagenumbers]{cvpr} 

\usepackage[most]{tcolorbox}
\usepackage{wrapfig}
\usepackage{multirow}
\usepackage{graphicx,xcolor,float}
\usepackage{subcaption}
\usepackage{threeparttable}
\usepackage[ruled,vlined]{algorithm2e}
\usepackage{colortbl}
\usepackage{color}
\usepackage{multirow}
\usepackage{tabularx}
\usepackage{float}
\usepackage{graphicx}
\usepackage{booktabs}
\usepackage{arydshln}
\usepackage{enumitem}
\usepackage{wrapfig}
\usepackage{caption}
\usepackage{graphicx}
\usepackage{makecell}
\usepackage{float} 
\usepackage{mathtools}
\usepackage{subcaption}
\usepackage{bbding}
\usepackage{makecell}
\usepackage{tabularx}
\usepackage{amssymb,mathrsfs,amsmath}
\usepackage{pifont}
\usepackage{xcolor}
\usepackage{bbm}

\usepackage{fontawesome5}

\usepackage{listings}

\newcommand{\ourmethod}{{\fontfamily{lmtt}\selectfont \textbf{LaMem-VLA}}\xspace}

\makeatletter
\newcommand{\thickhline}{%
    \noalign{\ifnum0=`}\fi\hrule height 1pt
    \futurelet\reserved@a\@xhline
}

\newcommand{\Rmnum}[1]{\expandafter\@slowromancap\romannumeral #1@}
\makeatother

\definecolor{myviolet}{RGB}{168,77,153}
\definecolor{myblue}{RGB}{60,118,185}
\definecolor{mygreen}{RGB}{70,145,80}
\definecolor{cvprblue}{rgb}{0.21,0.49,0.74}

\definecolor{cred}{HTML}{FF6B6B}
\definecolor{cyellow}{HTML}{FEC260}
\definecolor{cgreen}{HTML}{70AD47}
\definecolor{cblue}{HTML}{4D96FF}
\definecolor{cpurple}{HTML}{2A0944}
\definecolor{ggray}{RGB}{127,127,127}
\definecolor{aliceblue}{rgb}{0.94,0.97,1.0}
\definecolor{mygray}{gray}{.9}
\definecolor{mydarkblue}{rgb}{0,0.08,0.45}
\definecolor{chatgptgreen}{RGB}{16,163,127}

\newcommand{\obsbox}[1]{%
    \begin{tcolorbox}[
        colframe=black!70,
        colback=cyan!4,
        boxrule=1pt,
        arc=2mm,
        top=1pt,
        bottom=2pt,
        left=3pt,
        right=3pt,
        boxsep=1pt
    ]
        #1
    \end{tcolorbox}
}

\usepackage{bm}
\usepackage{microtype}

\definecolor{cvprblue}{rgb}{0.21,0.49,0.74}
\usepackage[pagebackref,breaklinks,colorlinks,allcolors=cvprblue]{hyperref}

\def\confName{CVPR}
\def\confYear{2027}

\title{Dual-Source Latent Memory Weaving for Vision-Language-Action Models}

\author{
Hongyu Qu$^{1}$ \quad
Jianzhe Gao$^{2}$ \quad
Xiaobin Hu$^{3}$ \quad
Shaohuan Yang$^{1}$ \quad
Xinlei Yu$^{3}$\\
Rui Yan$^{1}$ \quad
Wenguan Wang$^{2}$ \quad
Xiangbo Shu$^{1}$ \quad
Shuicheng Yan$^{3}$\\
{\small $^{1}$Nanjing University of Science and Technology
\quad $^{2}$Zhejiang University}\\
{\small $^{3}$National University of Singapore}
}

\begin{document}
\maketitle
\begin{abstract}
Mainstream Vision-Language-Action (VLA) models predict actions primarily from the current observation under a Markovian assumption, thus struggling with long-horizon, temporally dependent tasks. 
Existing memory-augmented VLAs either expand the observation window or retrieve history from the memory bank as auxiliary policy-side context.
However, they leave memory outside the native latent embedding space of VLA reasoning, preventing historical experience from being fluidly interleaved with multimodal reasoning and action formation.
To this end, we introduce \ourmethod, a latent-memory-native framework that reconstructs historical experience into latent memory tokens and directly interweaves them with VLA reasoning. 
At its core, \ourmethod introduces four coordinated components: 
\textit{\textbf{(i)}} a \textbf{curator} that organizes historical experience into two complementary visual and semantic memory vaults; \textit{\textbf{(ii)}} a \textbf{seeker} that queries both vaults using the multimodal cognition to retrieve context-relevant evidence; \textit{\textbf{(iii)}} a \textbf{condenser} that reconstructs the retrieved evidence into compact visual and semantic latent memory tokens; and \textit{\textbf{(iv)}} a \textbf{weaver} that injects these memory tokens with the current observation and instruction into one continuous embedding sequence.
By representing, retrieving, and consuming historical experience entirely in the same continuous latent space, \ourmethod enables memory to directly participate in VLA reasoning and guide action generation.
Experiments on SimplerEnv, LIBERO and  MIKASA-Robo demonstrate state-of-the-art success rates of 73.9\%, 97.6\%,
and 51.4\%, respectively.
In real-world
evaluation, \ourmethod ranks best on general and long-horizon suites.
\end{abstract}

    
\section{Introduction}
\label{sec:intro}
\begin{figure}[t]
    \centering
\includegraphics[width=\columnwidth]{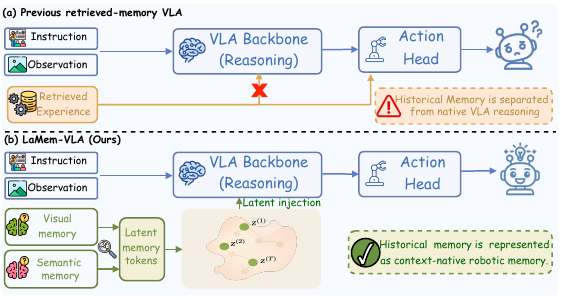}
 \vspace{-17pt}
    \caption{\small{
    Paradigm comparison of memory-augmented VLA models.
    (a) Unlike previous VLA models that store historical experience
    in an auxiliary memory bank and consume retrieved memory as
    external policy-side context,
    (b) \ourmethod treats historical experience as context-native
    latent memory, which is stored, retrieved, and consumed
    in the model embedding space.
    }}
    \vspace{-18pt}
    \label{fig:motivation}
\end{figure}
Vision-language-action (VLA) models~\cite{black2024pi_0,kim2025openvla,liu2025rdt,zhen20243d} have become a promising paradigm for general robotic manipulation.
By combining pretrained vision-language models~\cite{cheang2024gr,bai2023qwen,chen2024scaling} with policy learning~\cite{chi2025diffusion,zhang2025flowpolicy,octo_2023} on robotic data~\cite{o2024open,khazatsky2024droid,walke2023bridgedata,bu2025agibot}, they map visual observations and language instructions into executable action chunks.
Despite this progress, most existing VLA models~\cite{kim2025openvla,black2024pi_0,li2024cogact} implicitly rely on a Markovian assumption, predicting actions primarily from the current observation without considering temporal dependencies.
This simplification creates a \textit{temporal short-horizon bias}:
VLA models can react to the currently visible state, but struggle to reason over previous state transitions, completed operation steps, and the current phase of a multi-step task, making long-horizon manipulation particularly challenging.

Recent efforts have sought to alleviate temporal short-horizon bias by augmenting VLA models~\cite{chi2025diffusion,li2024language,ze20243d} with historical context or memory mechanisms along two axes.
\textit{\textbf{(i)}} One line of work incorporates short-horizon episode context by concatenating historical frames~\citep{li2024towards,fan2025interleave} or extending the input into a video sequence~\citep{li2026towards,koo2025hamlet,hu2026resolving}.
Although such designs expose recent state changes, they incur computational and memory costs that grow with the context length, while the fixed temporal horizon discards potentially relevant evidence outside the input window.
\textit{\textbf{(ii)}} Another line of work~\citep{shi2025memoryvla,sridhar2025memer,li2026global} retrieves past trajectories or relevant historical tokens from an external bank to condition downstream action policies.
Although these methods demonstrate the value of historical experience for long-horizon manipulation, they suffer from an architectural limitation: the historical memory is stored outside the model's native token space and consumed as auxiliary policy-side context after the VLA reasoning.
This separation prevents memory from being fluidly interleaved with the internal reasoning where the VLA model jointly perceives the scene, interprets the instruction, and resolves action queries before action formation.

This limitation raises a fundamental question for memory-dependent VLA models: \textit{whether historical experience can be represented as context-native robotic memory, stored, retrieved, and consumed in the same continuous space where the VLA model already perceives, reasons, and acts?}
Latent embedding space~\cite{wang2026monet,yang2026machine,li2025latent,bai2026latent,zhang2025memgen,yu2026vismem,yu2026latent} offers a natural answer to this question.
Modern VLA models already integrate visual observations and language instructions in a continuous token embedding space~\cite{zitkovich2023rt,kim2025openvla,black2024pi_0,li2024cogact}; therefore, robotic historical memory can be organized as machine-native latent memory tokens that are compatible with the internal reasoning process.
Under this formulation, robotic memory becomes part of the model's operating context rather than an auxiliary scaffold attached after multimodal reasoning.
Long-horizon robotic manipulation also calls for two complementary forms of historical memory: \textbf{visual memory} is visually dominant, preserving visually grounded evidence from the current episode, such as object locations and subtle state changes; \textbf{semantic memory} is semantically dominant, preserving task progress, contextual semantics, and action continuity across longer horizons.
Notably, their distinction lies in \textit{provenance and function} rather than in representation form: both are reconstructed as latent memory that can be consumed by the VLA model in the latent space.
This leads to our pivotal research question:
\vspace{-4pt}
\obsbox{
\includegraphics[height=0.9em]{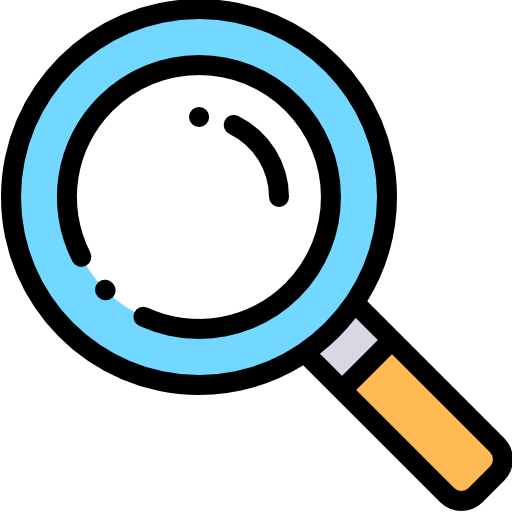}~
\textit{How can historical experience be organized as complementary visual and semantic memory, and reconstructed into compact latent memory tokens that interweave seamlessly with the VLA reasoning and action generation?}}
\vspace{-4pt}
To answer this question, we propose \ourmethod, a novel native latent memory framework for robotic manipulation, which explicitly organizes robotic history into two complementary memory vaults, and weaves dual-source memory into the model reasoning for memory-augmented action generation.
At its core, \ourmethod closes the loop between latent memory weaving and action reasoning through four coordinated modules:
First,  a \textbf{latent memory curator} factorizes past robotic experience into two complementary vaults: a visual memory vault for recent visual evidence and a semantic memory vault for semantic and action-continuity evidence.
Second, during the action reasoning process,  a \textbf{latent memory seeker} builds a context-aware query from the current multimodal cognition state (\ie, the visual and instruction tokens), and uses it to retrieve task-relevant historical evidence from dual memory vaults, grounding the present decision in past perceptual evidence and long-horizon semantic-action continuity. 
Third,  a \textbf{latent memory condenser} reconstructs the retrieved evidence as fixed-length visual and semantic latent memory tokens compatible with the VLA embedding space.
Finally,  a \textbf{latent memory weaver} inserts these condensed memory tokens into the action reasoning sequence before action query tokens are resolved, allowing historical memory to participate in the same latent reasoning process as the current image, instruction, and action queries.
The resulting memory-grounded action queries condition a diffusion-based action expert~\cite{chi2025diffusion,peebles2023scalable} to generate temporally aware robotic action sequences.

We conduct comprehensive evaluations on three simulation benchmarks (\ie, LIBERO~\cite{liu2023libero}, SimplerEnv-Bridge~\cite{li2025evaluating}, and MIKASA-Robo~\cite{cherepanov2026memory}) and six real-world manipulation tasks.
On LIBERO, SimplerEnv-Bridge, and MIKASA-Robo, \ourmethod achieves average success rates of \textbf{97.6}\%, \textbf{73.9}\%, and \textbf{51.4}\%, outperforming MemoryVLA~\cite{shi2025memoryvla} by \textbf{1.1}, \textbf{2.0}, and \textbf{10.2} points, respectively.
In real-world evaluations, \ourmethod achieves \textbf{77.3}\% and \textbf{61.3}\% average success on the Generalization and Long-horizon task suites, respectively, outperforming the strongest baseline by \textbf{15.3} points on the latter.
These results demonstrate the effectiveness of integrating complementary latent memory into VLA reasoning, particularly for long-horizon manipulation.
 
In summary, our main contributions are as follows:
\textbf{First}, we introduce a \textbf{new paradigm} for robotic memory in VLA models: historical experience is treated as context-native latent memory, which is stored, retrieved, and consumed in the model embedding space, supporting scene perception, instruction understanding, and action-intent formation within the same latent reasoning process.
\textbf{Second}, we propose \ourmethod, a \textbf{dual latent memory framework} for robotic manipulation, which explicitly organizes robotic history into two complementary memory vaults, and weaves dual-source memory into the model reasoning for action generation.
\textbf{Third}, we design a \textbf{latent memory condensing mechanism}, which transforms retrieved historical evidence into fixed-length latent visual and semantic tokens that are compatible with the VLA model embedding space.

\section{Related Work}
\label{sec:related}

\noindent\textbf{Vision-Language-Action (VLA) Models.}
Driven by advances in pretrained vision-language models (VLMs), VLA models~\citep{duan2025manipulate,huang2025rekep,black2024pi_0,liu2025hybridvla} have demonstrated promising performance in robotic control via mapping visual observations and language instructions to robot actions.
According to their action policy design, VLA models can be broadly classified into two paradigms: single-stream architecture and hierarchical architectures.
\textbf{(i)} Single-stream models~\citep{zhao2025cot,cen2025worldvla,wang2025unified,lin2025vote} directly discretize continuous actions~\citep{zitkovich2023rt,kim2025openvla} within a unified vision-language backbone, and autoregressively predict action tokens.
To meet the high control-frequency requirements in robotic manipulation, many works focus on inference efficiency optimization~\citep{wang2025spec,kim2025fine,zhang2026mole,wang2025bitvla}.
\textbf{(ii)} Hierarchical models~\citep{liu2025rdt,wen2025dexvla,wang2025vq} typically employ VLMs for high-level reasoning and planning, and adopt a separate generative policy, such as diffusion-based~\citep{black2024pi_0,li2024cogact} and flow-matching-based~\citep{deng2025graspvla,zhang2025flowpolicy} policies, to synthesize smooth and high-quality action trajectories. 
Despite these advances, these methods still predict actions primarily from the current observation and instruction without leveraging extended historical context, limiting their robustness in long-horizon manipulation tasks.
In contrast, our work aims to equip VLA models with explicit progress awareness via memory mechanisms, addressing memory-dependent tasks that current approaches fail to handle.

\begin{figure*}[!t]
    \centering
    \includegraphics[width=0.99\textwidth]{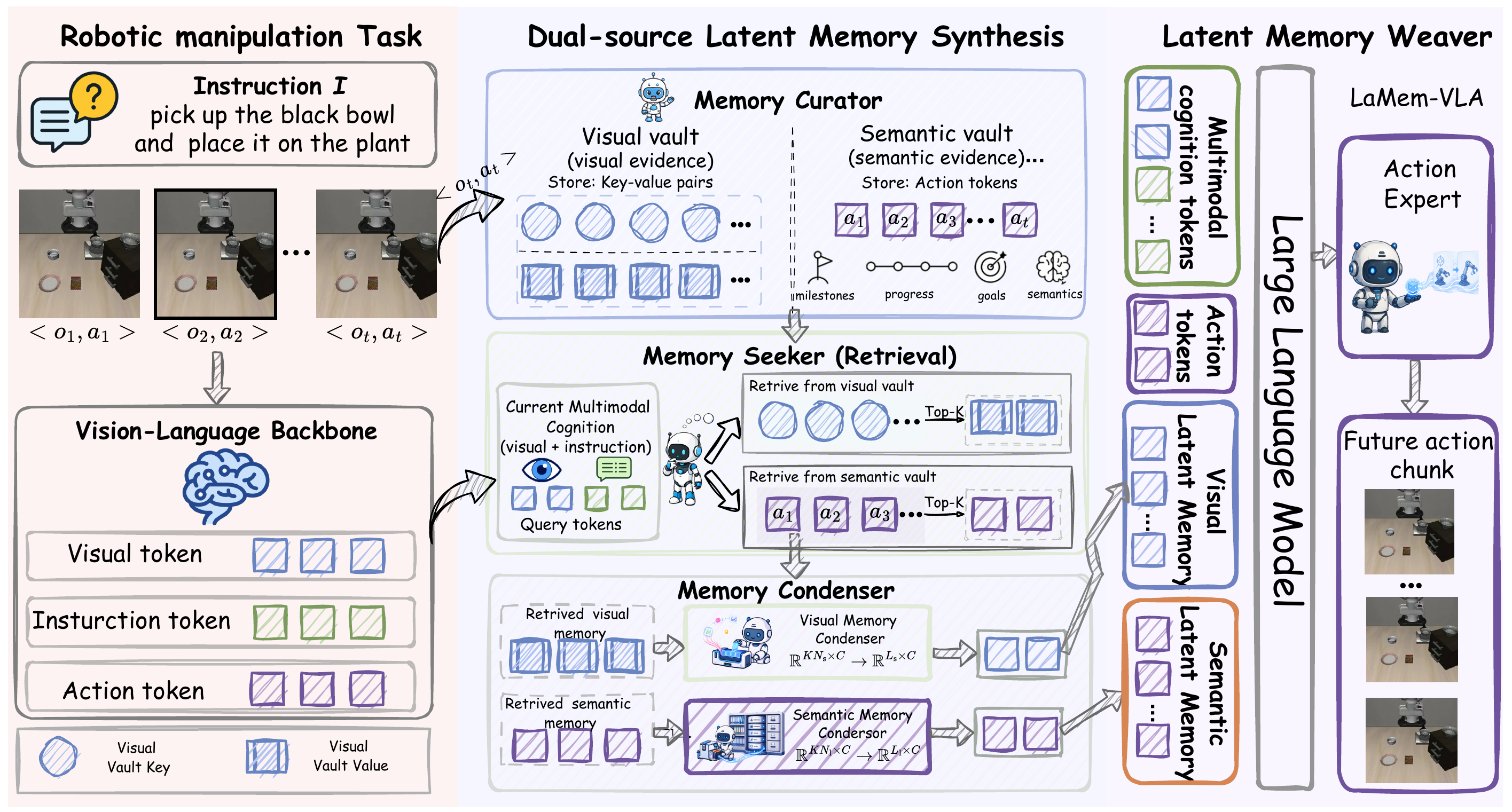}
    \vspace{-10pt}
    \caption{\small{\textbf{The Framework  of \ourmethod.}   
    The memory curator (\S\ref{sec:curator}) organizes visual tokens and action tokens into dual memory vaults, and the memory seeker (\S\ref{sec:seeker}) then retrieves task-relevant evidence from dual memory vaults based on visual and instruction tokens.
This retrieved evidence is compressed into fixed-length latent memory tokens by the memory condenser~(\S\ref{sec:seeker}). Finally, the memory weaver~(\S\ref{sec:weaver}) injects these latent memory tokens into the reasoning sequence before the action queries are resolved, producing memory-grounded action tokens that guide the action expert to generate future action chunks.
}}
    \label{fig:overview}
    \vspace{-12pt}
\end{figure*}

\noindent\textbf{Memory Mechanisms for Robotic Control.}
Mainstream VLA policies are formulated under a Markovian assumption, predicting actions primarily from the current observation and therefore lacking explicit awareness of task progress. 
Existing attempts to incorporate historical context can be broadly categorized into three groups. 
\textbf{(i)} \emph{Temporal-context expansion:} This line of work exposes the policy to extended observations by interleaving historical frames with language tokens~\citep{li2024towards,fan2025interleave} or aggregating multi-frame features~\citep{li2026towards,koo2025hamlet,hu2026resolving} within a predefined context window. 
However, their computational and memory costs grow with context length.
\textbf{(ii)} \emph{Sparse history abstraction:} Another paradigm compresses past interactions into lightweight proxy representations, such as recurrent latent states~\citep{li2026remem,li2024vision}, motion-centric cues~\citep{zheng2025tracevla}, action summaries~\citep{bu2025univla}, or a sparse set of representative observations~\citep{mark2026bpp}. However, such abstraction may discard fine-grained perceptual details and high-level semantic context.
\textbf{(iii)} \emph{External memory conditioning:}
A third direction~\citep{shi2025memoryvla,sridhar2025memer,li2026global} stores historical experience in auxiliary memory banks to support action prediction.
In particular, MemoryVLA~\citep{shi2025memoryvla} retrieves perceptual and cognitive history through cross-attention, gate-fuses it with current VLM features, and passes the resulting representations to a diffusion action expert.
In contrast, \ourmethod reconstructs retrieved history into compact context tokens that participate directly in the VLA backbone computation.
Specifically, query-conditioned condensers synthesize fixed-length visual and semantic memory tokens from selected historical entries.
The backbone then processes these tokens together with the current image, instruction, and action queries, allowing historical evidence to shape both multimodal contextualization and action-query representations.
The action expert subsequently decodes these memory-grounded representations into actions.
Thus, our distinction from MemoryVLA lies in the coordinated reconstruction and integration of historical evidence within the backbone, enabling memory to participate in action representation formation itself.

\section{Method}
\label{sec:method}

\subsection{Problem Formulation}
We formulate the robotic manipulation task in VLA models as a language-conditioned  Markovian decision-making problem. 
At each timestep $t$, the VLA policy~\cite{kim2025fine,kim2025openvla,li2024cogact}, $\bm{\Pi}_\theta$ takes  a natural-language instruction $\bm{I}$ and the current visual observation $\bm{o}_t$ as input, and predicts a chunk of future actions for executing the specified task:
\vspace{-4pt}
\begin{equation}
    \label{eq:problem_formulation}
         \bm{a}_{t:t+H-1} = \bm{\Pi}_\theta(\bm{o}_t, \bm{I}),
         \vspace{-4pt}
\end{equation}
where $H$ denotes the action horizon, and each action $a_t \in {\mathbb{R}}^{7}$ is a 7-DoF end-effector control vector, comprising a 3-DoF relative translation, a 3-DoF relative rotation, and a 1-DoF gripper state that jointly specify the robot's behavior.

\subsection{\ourmethod Model Architecture}
\ourmethod is an end-to-end native latent memory framework for robotic manipulation that directly weaves dual-source historical experience into the VLA reasoning to refine action generation, as shown in Fig.~\ref{fig:overview}.
\ourmethod closes the loop between latent memory reconstruction and action reasoning through four coordinated modules.
At each timestep, the vision-language backbone embeds current visual observation and instruction into the visual and instruction tokens, and learnable action queries are appended to the token sequence to obtain manipulation-relevant latent action representations.
First, a \textbf{latent memory curator} dynamically establishes and updates two complementary memory vaults: the visual memory vault $\mathcal{M}^\text{vis}$ stores visual tokens that preserve recent perceptual evidence, while the semantic memory vault $\mathcal{M}^\text{sem}$ stores action tokens that preserve semantic and action-continuity evidence across longer horizons.
Second, a \textbf{latent memory seeker} constructs a context-aware query $\bm{Q}_t^\text{con}$ from the current multimodal cognition representation and uses it to retrieve decision-relevant history from the two vaults.
Third, a \textbf{latent memory condenser} distills the retrieved raw memory content into compact latent visual memory tokens $\bm{M}^\text{vis}$ and latent semantic memory tokens $\bm{M}^\text{sem}$, respectively.
Finally, a \textbf{latent memory weaver} interleaves these tokens with the current image, instruction, and action queries to form a memory-augmented VLA memory-grounded representations. These tokens are then fed into a diffusion-based action expert  to generate action sequences.


\subsection{Latent Memory Curator}
\label{sec:curator}
The latent memory curator maintains the historical evidence that will later be retrieved, condensed, and woven into the VLA reasoning sequence.
\ourmethod adopts a 7B-parameter Prismatic vision-language model~\cite{karamcheti2024prismatic} as the backbone, which is further pretrained on the large-scale cross-embodiment real robotic dataset Open-X Embodiment~\cite{o2024open}.
For the current RGB observation, the backbone first extracts visual tokens with its vision encoder~\cite{oquabdinov2,zhai2023sigmoid} and projects them into the language embedding space to obtain final visual tokens $\bm{X}_t\!\in\!\mathbb{R}^{N_\text{i}\times C}$, where $N_\text{i}$ is the sequence length and $C$ is the hidden dimension.
These visual tokens are concatenated with the tokenized instruction and processed by LLaMA-7B~\cite{touvron2023llama}.
We append learnable action queries $\bm{Q}^\text{action}\!\in\!\mathbb{R}^{N_\text{a}\times C}$ to the sequence, and use their output hidden states $\bm{H}^\text{action}$ as compact action representations for downstream action prediction.
The curator factorizes the historical evidence into two memory vaults: a visual memory vault $\mathcal{M}^\text{vis}$ and a semantic memory vault $\mathcal{M}^\text{sem}$.
Both vaults are initialized as empty at the beginning of each episode.
Each vault has a maximum capacity of $L$ memory units.

\noindent\textbf{Visual Memory Vault.}
The visual memory vault $\mathcal{M}^\text{vis}$ stores visual perceptual evidence from the current episode.
Each visual memory unit is represented as a key-value pair $\bm{m}_\text{s}^i=(\bm{k}_\text{s},\bm{v}_\text{s})$:
the key $\bm{k}_\text{s}$ provides a concise retrieval summary of visual evidence, while the value $\bm{v}_\text{s}$ stores the latent visual memory content.
Specifically, at each timestep, a learnable compression module distills the current visual tokens $\bm{X}_t$ into a compact set of visual memory tokens, and their mean-pooled representation is used as the retrieval key:
\vspace{-4pt}
\begin{equation}
\label{eq:visual_memory_bank}
    \begin{aligned}
        \bm{v}_\text{s} &= \mathcal{C}_\text{s}(\bm{X}_t) \in \mathbb{R}^{N_\text{s} \times C}, &
        \bm{k}_\text{s} &= \mathrm{MeanPool}(\bm{v}_\text{s}) \in \mathbb{R}^{C},
    \end{aligned}
    \vspace{-2pt}
\end{equation}
where $\mathcal{C}_\text{s}$ is an SE-bottleneck compression module~\cite{hu2018squeeze}, and the unit $\bm{m}_\text{s}$ is then appended to $\mathcal{M}^\text{vis}$.

\noindent\textbf{Semantic Memory Vault.}
The semantic memory vault $\mathcal{M}^\text{sem}$ stores the action hidden states that track task progress and action continuity across long horizons.
At each timestep, the curator directly writes the action hidden states $\bm{H}^\text{action}$  into the vault as one semantic memory unit $\bm{m}_\text{l}$. 

\noindent\textbf{Memory Vault Updating Strategy.}
After a new memory unit is written into its corresponding vault, the memory curator applies a compression strategy only when the number of stored units exceeds the capacity $L$.
For the visual stream, let $\mathcal{M}^\text{vis}=\{\bm{m}_\text{s}^i=(\bm{k}_\text{s}^i,\bm{v}_\text{s}^i)\}_{i=1}^{n_\text{s}}$ after insertion.
If $n_\text{s}>L$, we compute the cosine similarity between temporally adjacent keys and select the most redundant adjacent pair:
\vspace{-4pt}
\begin{equation}
    \label{eq:short_memory_capacity}
    i_\text{s}^* =
    \arg\max_{1 \le i < n_\text{s}}
    \mathrm{cos}(\bm{k}_\text{s}^i,\bm{k}_\text{s}^{i+1}).
    \vspace{-4pt}
\end{equation}
The selected pair is consolidated by averaging both its key and value tokens:
\vspace{-4pt}
\begin{equation}
\label{eq:short_memory_merge}
\begin{alignedat}{3}
\tilde{\bm{k}}_\text{s} &=
\frac{1}{2}(\bm{k}_\text{s}^{i_\text{s}^*}+\bm{k}_\text{s}^{i_\text{s}^*+1}),
\;&
\tilde{\bm{v}}_\text{s} &=
\frac{1}{2}(\bm{v}_\text{s}^{i_\text{s}^*}+\bm{v}_\text{s}^{i_\text{s}^*+1}),
\;&
\tilde{\bm{m}}_\text{s} &=
(\tilde{\bm{k}}_\text{s},\tilde{\bm{v}}_\text{s}).
\end{alignedat}
\vspace{-2pt}
\end{equation}
The two adjacent units are replaced by $\tilde{\bm{m}}_\text{s}$, reducing redundancy while preserving their shared visual evidence. For the semantic vault, adjacent units are ranked by the similarity of their mean-pooled action representations and consolidated by averaging. More details are provided in Appendix.

\subsection{Learning to synthesize latent memory with Dual-source Vault}
\label{sec:seeker}
\ourmethod does not expose memory vaults to the action policy as raw auxiliary context.
Instead, it treats the two vaults as latent evidence substrates: 
given the current multimodal cognition representation, \ourmethod first retrieves relevant evidence from the two vaults and then reconstructs it into compact memory tokens that are native to the VLA embedding space.
This design decouples memory storage from memory consumption, allowing historical experience to remain flexible in the vaults while entering action reasoning through a bounded latent interface.

\noindent\textbf{Latent Memory Seeker.}
The latent memory seeker retrieves evidence from the memory vaults according to the current multimodal context rather than the visual observation or language instruction alone.
Given the current visual tokens $\bm{X}_t$ and instruction tokens $\bm{I}$, the VLA backbone produces a context-aware query $\bm{Q}_t^\text{con}$ that encodes the current visual-linguistic cognition. The seeker then appends learnable query slots $\bm{Q}^\text{init}\in\mathbb{R}^{K_\text{q}\times C}$ to $\bm{Q}_t^\text{con}$ and only updates the query slots  with a lightweight query builder:
\vspace{-4pt}
\begin{equation}
    \label{eq:latent_memory_query}
    \bm{Q}_t = \mathcal{B}([\bm{Q}_t^\text{con}; \bm{Q}^\text{init}])[-K_\text{q}:]
    \in \mathbb{R}^{K_\text{q}\times C},
    \vspace{-4pt}
\end{equation}
where $\mathcal{B}$ is a transformer-based query builder (more details in Appendix) and $\bm{Q}_t$ serves as the shared memory hook for both vaults.
We apply masked attention inside $\mathcal{B}$ such that the appended query slots read from $\bm{Q}_t^\text{con}$, while the original multimodal hidden states are not perturbed by the query slots.
The mean-pooled query
$\bm{q}_t\!=\!\mathrm{MeanPool}(\bm{Q}_t)\in\mathbb{R}^{C}$
is used as the global retrieval vector.

The memory seeker then uses $\bm{q}_t$ to retrieve context-relevant units from the visual vault $\mathcal{M}^\text{vis}=\{(\bm{k}_\text{s}^i,\bm{v}_\text{s}^i)\}_{i=1}^{|\mathcal{M}^\text{vis}|}$ by cosine similarity:
\vspace{-4pt}
\begin{equation}
    \label{eq:short_memory_retrieval}
    \begin{aligned}
        &\mathcal{I}_\text{s}
        = \mathrm{Top\!-\!}K
        \left(\{\mathrm{cos}(\bm{q}_t,\bm{k}_\text{s}^i)\}_{i=1}^{|\mathcal{M}^\text{vis}|}\right),\\
        &\bm{Z}^\text{vis}
        = \mathrm{Concat}_{i\in\mathcal{I}_\text{s}}(\bm{v}_\text{s}^i)
        \in \mathbb{R}^{KN_\text{s}\times C}.
    \end{aligned}
    \vspace{-4pt}
\end{equation}
where $K$ is the number of retrieved visual memory units.
For semantic memory retrieval, the seeker similarly ranks the units in $\mathcal{M}^\text{sem}=\{\bm{m}_\text{l}^i\}_{i=1}^{|\mathcal{M}^\text{sem}|}$ via comparing the mean-pooled semantic units with the mean-pooled query $\bm{q}_t$.
It then selects the $\mathrm{Top\!-\!}K$ semantic memory units and stacks them into $\bm{Z}^\text{sem}\in \mathbb{R}^{KN_\text{l}\times C}$.
The retrieved sets $\bm{Z}^\text{vis}$ and $\bm{Z}^\text{sem}$ are passed to the memory condenser below as visual evidence and semantic evidence, respectively, instead of being inserted verbatim into the VLA sequence.

\noindent\textbf{Latent Memory Condenser.}
Directly inserting the retrieved evidence sequences into the VLA sequence would expand the reasoning context and introduce redundant historical tokens, so the latent memory condenser reconstructs them as fixed-length tokens and maps them into the VLA reasoning embedding space.
Specifically, we introduce learnable visual memory slots $\bm{T}_\text{s}\in\mathbb{R}^{L_\text{s}\times C}$ and semantic memory slots $\bm{T}_\text{l}\in\mathbb{R}^{L_\text{l} \times C}$, and update them with lightweight memory formers conditioned on the context query tokens $\bm{Q}_t\in\mathbb{R}^{K_\text{q}\times C}$ and the retrieved evidence:
\vspace{-4pt}
\begin{equation}
    \label{eq:memory_condensation}
    \begin{aligned}
        &\bm{M}^\text{vis} =
        \mathcal{F}_\text{v}([\bm{Q}_t; \bm{Z}^\text{vis}; \bm{T}_\text{s}])[-L_\text{s}:],\\
        &\bm{M}^\text{sem} =
        \mathcal{F}_\text{c}([\bm{Q}_t; \bm{Z}^\text{sem}; \bm{T}_\text{l}])[-L_\text{l}:].
    \end{aligned}
    \vspace{-4pt}
\end{equation}
where $\mathcal{F}_\text{v}$ and $\mathcal{F}_\text{c}$ denote lightweight transformer-style memory formers similar to $\mathcal{B}$ for the visual and semantic memory, respectively.
The resulting $\bm{M}^\text{vis}$ and $\bm{M}^\text{sem}$ are query-conditioned latent visual and semantic memory tokens in the same $C$-dimensional embedding space used by VLA reasoning.
This fixed-length property makes the injected memory independent of the retrieval size while preserving query-relevant evidence from the two complementary vaults.

\begin{table*}[t]
\centering
\caption{
\small\textbf{Quantitative comparison results on MIKASA-Robo~\citep{cherepanov2026memory} (\S\ref{sec:mikasa})} with the Franka robot.
}
\vspace{-7pt}
\label{tab:mikasa}
\resizebox{\linewidth}{!}{
\setlength\tabcolsep{12pt}
\renewcommand\arraystretch{1.0}
\begin{tabular}{l|c|ccccc|c}
\thickhline
\rowcolor{mygray}
Method
& Publication
& \makecell{ShellGame\\Touch}
& \makecell{Intercept\\Medium}
& \makecell{Remb.\\Color-3}
& \makecell{Remb.\\Color-5}
& \makecell{Remb.\\Color-9}
& \makecell{Avg.\\Success} \\
\hline
\hline 
\rowcolor{white}
CronusVLA~\citep{li2026towards}
& AAAI2026
& 32 & 5 & 31 & 13 & 9 & 18.0 \\

SpatialVLA~\citep{qu2025spatialvla}
& RSS2025
& 23 & 27 & 27 & 17 & 11 & 21.0 \\

OpenVLA-OFT~\citep{kim2025fine}
& ArXiv2025
& 47 & 14 & \textbf{59} & 16 & 6 & 28.4 \\

$\pi_0$~\citep{black2024pi_0}
& RSS2025
& 33 & 42 & 35 & 22 & 15 & 29.4 \\

Octo~\citep{octo_2023}
& RSS2024
& 46 & 39 & 45 & 17 & 11 & 31.6 \\

MemoryVLA~\citep{shi2025memoryvla}
& ICLR2026
& 88 & 24 & 44 & 30 & 20 & 41.2 \\
\hline
\ourmethod~\textbf{(Ours)}
& --
& \textbf{94}
& \textbf{43}
& 58
& \textbf{37}
& \textbf{25}
& \textbf{51.4} \\
\hline
\end{tabular}
}
\vspace{-6pt}
\end{table*}
\begin{table}[t]
\centering
\caption{
\small\textbf{Quantitative comparison results on SimplerEnv-Bridge~\cite{li2025evaluating}} (\S\ref{sec:simpler}) with WidowX robot.}
\vspace{-7pt}
\label{tab:simplerenv_bridge}
\resizebox{\linewidth}{!}{
\setlength\tabcolsep{2pt}
\renewcommand\arraystretch{1.0}
\begin{tabular}{l|c|cccc|c}
\thickhline
\rowcolor{mygray}
Method
& Publication
& \makecell{Spoon\\on Towel}
& \makecell{Carrot\\on Plate}
& \makecell{Stack\\Cube}
& \makecell{Eggplant\\in Basket}
& \makecell{Avg.\\Success} \\
\hline 
\hline
\rowcolor{white}
RT-1-X~\citep{o2024open}              & ICRA2024 & 0.0  & 4.2  & 0.0  & 0.0   & 1.1  \\
OpenVLA~\citep{kim2025openvla}             & CoRL2025 & 4.2  & 0.0  & 0.0  & 12.5  & 4.2  \\
TraceVLA~\citep{zheng2025tracevla}         & ICLR2025 & 12.5 & 16.6 & 16.6 & 65.0  & 27.7 \\
SpatialVLA~\citep{qu2025spatialvla}        & RSS2025  & 16.7 & 25.0 & 29.2 & 100.0 & 42.7 \\
CogACT~\citep{li2024cogact}          & ArXiv2024 & 58.3 & 45.8 & 29.2 & 95.8  & 57.3 \\
$\pi_0$~\citep{black2024pi_0}          & RSS2025& 83.8 & 52.5 & 52.5 & 87.9 & 69.2 \\
DreamVLA~\citep{zhang2026dreamvla}          & NeurIPS2025&45.8 &45.8 &25.0 &87.5 &51.0 \\
ThinkAct~\citep{huang2026thinkact}          & NeurIPS2025& 58.3& 37.5& 8.7 &70.8 &43.8 \\
CronusVLA~\citep{li2026towards}          & AAAI2026& 66.7& 54.2& 20.8& 100.0& 60.4 \\
MemoryVLA~\citep{shi2025memoryvla}         & ICLR2026 & 75.0 & 75.0 & 37.5  & 100.0 & 71.9 \\
SemanticVLA~\citep{ni2026semanticvla}          & CVPR2026& 83.6 &54.5& 40.3& 81.8 &65.1\\
\hline 
\ourmethod~\textbf{(Ours)}
& --
& 83.3 & 75.0 & 41.7 & 95.8
& \textbf{73.9}\\
\hline
\end{tabular}
}
\vspace{-14pt}
\end{table}

\subsection{Interweaving Memory into Action Reasoning in Latent Space}
\label{sec:weaver}
\noindent\textbf{Latent Memory Weaver for Guidance Generation.}
Rather than passing the condensed memory tokens to a policy head as external conditioning alone, the latent memory weaver injects the synthesized memory into the VLA reasoning sequence before action queries are resolved during the current policy step.
Although the two types of latent memory differ in provenance, they share the same latent token interface.
Given the condensed visual memory tokens $\bm{M}^\text{vis}$ and semantic memory tokens $\bm{M}^\text{sem}$, the weaver constructs the memory-augmented VLA input sequence $\bm{S}_t$:
\vspace{-4pt}
\begin{equation}
    \label{eq:memory_weaver}
    \begin{gathered}
        \bm{S}_t =
        [\bm{M}^\text{vis} + \mathbf{1}_{L_\text{s}}\bm{b}_\text{s}^\top;
        \bm{M}^\text{sem} + \mathbf{1}_{L_\text{l}}\bm{b}_\text{l}^\top;
        \bm{X}_t; \bm{I};
        \bm{Q}^\text{action}],\\
        \bm{Z}^\text{action} =
        \mathrm{VLM}(\bm{S}_t)[-N_\text{a}:].
    \end{gathered}
    \vspace{-1pt}
\end{equation}
where $\bm{b}_\text{s},\bm{b}_\text{l}\in\mathbb{R}^{C}$ are learnable source embeddings for the two memory streams, and $\mathbf{1}_{L_\text{s}},\mathbf{1}_{L_\text{l}}$ are all-one column vectors that broadcast them over the visual and semantic memory tokens according to their respective lengths.
Because the two sources of memory tokens are part of the model input sequence, they participate in self-attention with the current observation, language instruction, and action queries.

\noindent\textbf{Diffusion-based Action Expert.}
After the vision-language backbone produces the memory-grounded action tokens $\bm{Z}^\text{action}$, the diffusion-based action expert decodes them into a continuous action chunk.
Following the diffusion policy~\cite{chi2025diffusion}, we formulate action generation as conditional denoising over an action chunk.
Let $\bm{a}_{t:t+H-1}^{0}$ denote the clean expert action chunk and $\bm{a}_{t:t+H-1}^{n}$ denote its noisy version at diffusion step $n$.
At each denoising step, the noisy action tokens are injected with the diffusion timestep embedding and conditioned on action tokens $\bm{Z}^\text{action}$.

The diffusion expert $\epsilon_\theta$ is trained with mean squared error (MSE) loss to predict the injected noise under these action and memory conditions:
\vspace{-4pt}
\begin{equation}
    \label{eq:diffusion_loss}
    \mathcal{L}_\text{action}
    =
    \mathbb{E}_{n,\epsilon}
    \left[
    \left\|
    \epsilon -
        \epsilon_\theta(\bm{a}_{t:t+H-1}^{n}, n, \bm{Z}^\text{action})
    \right\|_2^2
    \right].
    \vspace{-4pt}
\end{equation}
During inference, DDIM sampling~\cite{songdenoising} iteratively denoises the action chunk under the same conditions, producing history-aware continuous $7$-DoF control actions.
\begin{table*}[t]
\centering
\caption{
\small\textbf{Quantitative comparison results on LIBERO~\citep{liu2023libero} (\S\ref{sec:libero}) with the Franka robot.}
Success rates (\%) are reported across five suites.
$^{*}$ indicates methods using additional proprioceptive and wrist-camera inputs.
For methods without Long-90 results, we report the average over the first four suites.
}
\vspace{-8pt}
\label{tab:libero}
\resizebox{\linewidth}{!}{
\setlength\tabcolsep{14pt}
\renewcommand\arraystretch{1.0}
\begin{tabular}{l|c|ccccc|c}
\thickhline
\rowcolor{mygray}
Method & Publication & Spatial & Object & Goal & Long-10 & Long-90
& \makecell{Avg. Success} \\
\hline
\hline
\rowcolor{white}
Diffusion Policy~\citep{chi2025diffusion}
& IJRR2025
& 78.3 & 92.5 & 68.3 & 50.5 & -- & 72.4 \\

Octo~\citep{octo_2023}
& RSS2024
& 78.9 & 85.7 & 84.6 & 51.1 & -- & 75.1 \\

SpatialVLA~\citep{qu2025spatialvla}
& RSS2025
& 88.2 & 89.9 & 78.6 & 55.5 & 46.2 & 71.7 \\

OpenVLA~\citep{kim2025openvla}
& CoRL2025
& 84.7 & 88.4 & 79.2 & 53.7 & 73.5 & 75.9 \\

CoT-VLA~\citep{zhao2025cot}
& CVPR2025
& 87.5 & 91.6 & 87.6 & 69.0 & -- & 83.9 \\

$\pi_0$-FAST$^{*}$~\citep{pertsch2025fast}
& ArXiv2025
& 96.4 & 96.8 & 88.6 & 60.2 & 83.1 & 85.0 \\

CogACT~\citep{li2024cogact}
& ArXiv2024
& 97.2 & 98.0 & 90.2 & 88.8 & 92.1 & 93.2 \\

$\pi_0^{*}$~\citep{black2024pi_0}
& RSS2025
& 96.8 & 98.8 & 95.8 & 85.2 & -- & 94.2 \\

TraceVLA~\citep{zheng2025tracevla}
& ICLR2025
& 84.6 & 85.2 & 75.1 & 54.1 & -- & 74.8 \\

MemoryVLA~\citep{shi2025memoryvla}
& ICLR2026
& 98.4 & 98.4 & 96.4 & 93.4 & 95.6 & 96.5 \\

Fast-ThinkAct~\citep{huang2026fast}
& CVPR2026
& 92.0 & 97.2 & 90.2 & 79.4 & -- & 89.7 \\

SemanticVLA~\citep{ni2026semanticvla}
& CVPR2026
& 98.0 & 98.6 & 96.8 & 94.4 & -- & 97.0 \\

LARA~\citep{liu2026lara}
& ICML2026
& 96.5 & 97.5 & 96.0 & 92.5 & -- & 95.6 \\

TTF-VLA~\citep{liu2026ttf}
& AAAI2026
& 84.0 & 73.0 & 81.0 & 58.0 & -- & 74.0 \\

CronusVLA~\citep{li2026towards}
& AAAI2026
& 97.3 & \textbf{99.6} & 96.9 & 94.0 & -- & 97.0 \\
\hline
\ourmethod~\textbf{(Ours)}
& --
& \textbf{98.8} & 99.0 & \textbf{97.2}
& \textbf{95.8} & \textbf{97.0} & \textbf{97.6} \\
\hline
\end{tabular}
}
\vspace{-8pt}
\end{table*}

\section{Experiment}
\label{sec:exper}

\subsection{Implementation Details}
We instantiate \ourmethod with a 7B-parameter Prismatic VLM backbone and a diffusion action expert with approximately 300M parameters.
We train the model on 8 NVIDIA H800 GPUs using PyTorch FSDP.
Each GPU processes 32 samples, resulting in a global batch size of 256, and the learning rate is set to $2\times10^{-5}$.
Both the memory vaults keep a maximum capacity of $L\!=\!16$ units.
The seeker retrieves the top $K\!=\!8$ units from each vault, and the condenser reconstructs the retrieved evidence into $L_\text{s}\!=\!8$ visual memory tokens and $L_\text{l}\!=\!4$ semantic memory tokens.
The query builder $\mathcal{B}$ and the two memory formers $\mathcal{F}_\text{v}$ and $\mathcal{F}_\text{c}$ are implemented as two-layer transformer blocks with masked attention (more details in Appendix).
During inference, the diffusion action expert generates actions with DDIM~\cite{songdenoising} sampling using 10 denoising steps. Both \ourmethod and MemoryVLA~\citep{shi2025memoryvla} are fine-tuned separately on each simulation dataset following the same protocol.
Details of temporal sample organization, and memory construction and update procedures are provided in
Appendix.

\subsection{Simulated  Evaluation on Mikasa-Robo}
\label{sec:mikasa}
\noindent\textbf{Training and Evaluation Protocol.}
We evaluate long-horizon temporal tasks on Mikasa-Robo~\citep{cherepanov2026memory} and compare it with representative temporal-context VLAs (\ie, CronusVLA~\cite{li2025cronusvla} and MemoryVLA~\cite{shi2025memoryvla}) that explicitly exploit
historical observations or memory.
Following the standard protocol, we use 250 demonstrations per task, $128\times128$ RGB observations, end-effector control, and 100 evaluation episodes per task.
The five tasks are jointly trained for 20K steps, with validation performed every 1K steps, and results are reported using the best validation checkpoint.

\noindent\textbf{Evaluation Results.}
\ourmethod consistently achieves the highest performance on four of the five tasks, with a \textbf{10.2}\% improvement in average success rate over the previous state of the art, including a \textbf{7.0}\% gain on RememberColor-5. The results are shown in Table~\ref{tab:mikasa}.

\subsection{Simulated  Evaluation on SimplerEnv-Bridge}
\label{sec:simpler}
\noindent\textbf{Training and Evaluation Protocol.}
SimplerEnv-Bridge~\cite{li2025evaluating} evaluates real-to-sim generalization of robot manipulation policies trained on real-world data.
\ourmethod is trained on the Bridge v2 dataset~\cite{walke2023bridgedata} for 50k optimization steps, with validation performed every 2.5k steps.
The final results are reported using the checkpoint that achieves the best validation performance. For evaluation, each task is tested over 24 trials, and we report the average success rate across trials. TraceVLA~\cite{zheng2025tracevla}, CronusVLA~\cite{li2025cronusvla}, and MemoryVLA~\cite{shi2025memoryvla} are representative temporal-context VLAs that explicitly exploit historical observations or memory.

\noindent\textbf{Evaluation Results.}
As shown in Table~\ref{tab:simplerenv_bridge}, \ourmethod achieves the highest average success rate of \textbf{73.9}\% on SimplerEnv-Bridge, outperforming CogACT~\cite{li2024cogact} by \textbf{16.6} points and surpassing recent VLAs such as $\pi_0$~\cite{black2024pi_0} and SemanticVLA~\citep{ni2026semanticvla}. These results indicate that injecting dual latent memory tokens into the reasoning sequence provides effective historical context for action prediction. 



\begin{figure*}[!t]
    \centering
    \includegraphics[width=0.99\textwidth]{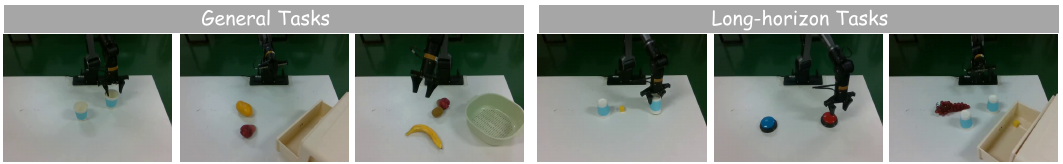}
    \vspace{-12pt}
\includegraphics[width=0.99\textwidth]{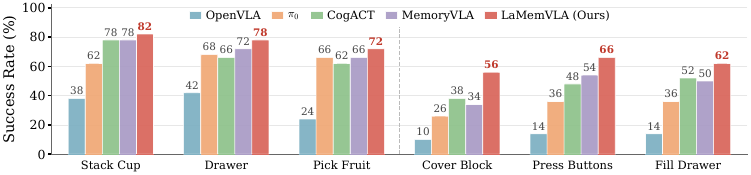}
    \vspace{-0pt}
    \caption{\small{
    \textbf{Real-world task setup and evaluation results} (\S\ref{sec:real}).
    We evaluate \ourmethod against leading models~\citep{
    kim2025openvla,
    black2024pi_0,
    li2024cogact,
    shi2025memoryvla}
    on the Generalization and Long-horizon task suites.}
    }
    \label{fig:realworld}
    \vspace{-13pt}
\end{figure*}

\subsection{Simulated  Evaluation on LIBERO}
\label{sec:libero}
\noindent\textbf{Training and Evaluation Protocol.}
We evaluate \ourmethod on the LIBERO~\cite{liu2023libero} using a Franka robot across five suites: spatial awareness (Spatial), object manipulation (Object), goal completion (Goal), and long-horizon reasoning (Long-10 and Long-90). The first four suites comprise 10 tasks each, while Long-90 contains 90 tasks. Following the OpenVLA protocol~\cite{kim2025openvla}, we use 50 demonstrations per task. We train separate models for Spatial, Object, and Goal for 20k optimization steps, and jointly train on Long-10 and Long-90 for 40k steps. We report success rates for each suite and the overall average over 50 rollouts per task. TraceVLA~\cite{zheng2025tracevla}, CronusVLA~\cite{li2025cronusvla}, TTF-VLA~\cite{liu2026ttf}, and MemoryVLA~\cite{shi2025memoryvla} are representative temporal-context VLAs that explicitly exploit historical observations or memory for temporally dependent manipulation tasks.

\noindent\textbf{Evaluation Results.}
As shown in Table~\ref{tab:libero}, \ourmethod achieves the highest average success rate of \textbf{97.6}\% across the five LIBERO suites, outperforming MemoryVLA~\citep{shi2025memoryvla} and CogACT~\citep{li2024cogact} by \textbf{1.1} and \textbf{4.4} points, respectively. It also surpasses $\pi_0$~\citep{black2024pi_0} by \textbf{3.5} points over the first four suites, without using the additional proprioceptive and wrist-camera inputs. 
Notably, \ourmethod obtains \textbf{95.8}\% on Long-10 and \textbf{97.0}\% on Long-90, exceeding MemoryVLA by \textbf{2.4} and \textbf{1.4} points, highlighting its long-horizon manipulation capabilities. 

\begin{table*}[t]
\begin{minipage}[t]{0.49\linewidth}
\centering
\caption{
\small \textbf{Detailed analysis of dual-source latent memory} on Mikasa-Robo~\cite{cherepanov2026memory} and LIBERO~\cite{liu2023libero} (\S\ref{sec:abation}).
}
\vspace{-7pt}
\label{tab:ablation_dual_memory}
\resizebox{\linewidth}{!}{
\setlength\tabcolsep{10pt}
\renewcommand\arraystretch{1.0}
\begin{tabular}{l|cc}
\thickhline
\rowcolor{mygray}
Method
& Mikasa-Robo
& Long-90 \\
\hline
\hline
w/o Dual-source Memory       & 29.2 & 92.1 \\
w/o Visual Memory  & 41.8 & 95.4  \\ 
w/o Semantic Memory & 43.6 & 94.8 \\\hline
\ourmethod~\textbf{(Ours)}
& \textbf{51.4} & \textbf{97.0} \\\hline
\end{tabular}
}

\end{minipage}
\hfill
\begin{minipage}[t]{0.5\linewidth}
\centering
\caption{\small \textbf{Ablation study of latent-native memory integration} Mikasa-Robo~\cite{cherepanov2026memory} and LIBERO~\citep{liu2023libero} (\S\ref{sec:abation}).
}
\vspace{-7pt}
\label{tab:ablation_latent_native_memory}
\resizebox{\linewidth}{!}{
\setlength\tabcolsep{8pt}
\renewcommand\arraystretch{1.0}
\begin{tabular}{l|cc}
\thickhline
\rowcolor{mygray}
Method
& Mikasa-Robo
& Long-90 \\
\hline
\hline
\textsc{Baseline}       & 29.2 & 92.1 \\
Policy-side Memory
& 41.2 & 94.8 \\
Raw Retrieval Conditioning
& 40.8 & 95.1 \\
\hline
\ourmethod~\textbf{(Ours)}
& \textbf{51.4} & \textbf{97.0} \\
\hline
\end{tabular}
}

\end{minipage}
\vspace{-7pt}
\end{table*}
\begin{figure*}[t]
\centering
\begin{minipage}[t]{0.48\linewidth}
    \centering
    \includegraphics[width=\linewidth]{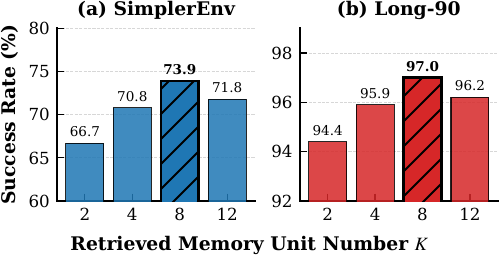}
    \vspace{-20pt}
    \caption{\small \textbf{Ablation of retrieved memory unit number $K$} on SimplerEnv~\cite{li2025evaluating} and Long-90~\cite{liu2023libero} (\S\ref{sec:abation}).}
    \label{tab:ablation_retrieved_k}
\end{minipage}
\vspace{-6pt}
\hfill
\begin{minipage}[t]{0.51\linewidth}
    \centering   \includegraphics[width=\linewidth]{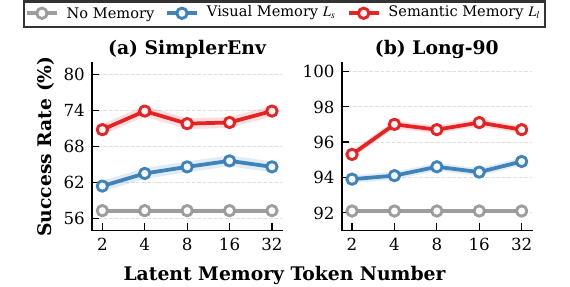}
    \vspace{-19pt}
    \caption{\small \textbf{Ablation study of the latent memory token number} on SimplerEnv~\cite{li2025evaluating} and Long-90~\cite{liu2023libero} (\S\ref{sec:abation}).}
 \label{fig:latentnum}
\end{minipage}
\vspace{-7pt}
\end{figure*}

\subsection{Real-world Evaluation}
\label{sec:real}
\noindent\textbf{Training and Evaluation Protocol.}
To validate the effectiveness of \ourmethod in real-world robotic manipulation, we conduct experiments on the AgileX Piper, a 6-DoF single-arm manipulator equipped with a parallel-jaw gripper and a single exocentric Intel RealSense D435i RGB camera. Our evaluation protocol encompasses two task suites:  \textit{General Tasks} and \textit{Long-horizon Tasks}. General Tasks assess manipulation from current observations, and Long-horizon Tasks require historical context to resolve ambiguities during task execution. For the former, we collect 200 expert demonstrations per task and conduct $50$ evaluation rollouts. For the latter, we utilize 300  expert demonstrations and conduct $50$ rollouts per task. In all evaluations, object positions are randomly initialized. More details are provided in Appendix.

\noindent\textbf{Results on Real-World.}
As illustrated in Fig.~\ref{fig:realworld}, \ourmethod achieves average success rates of 77.3\% and 61.3\%  on General and Long-horizon Tasks, respectively. On the general tasks, it exceeds the strongest baseline on every task, with a gain of up to 6 points on Pick Fruit. 
On the long-horizon tasks, where correct decisions depend on earlier events, \ourmethod outperforms the strongest baseline by 15.3 points.
These results demonstrate strong real-world competence of \ourmethod. 
\subsection{Ablation Study}
\label{sec:abation}
\noindent\textbf{Effectiveness of Dual-source Latent Memory.}
We evaluate the contribution of each memory source by selectively removing it from the latent VLA input sequence.
We consider four settings: the full \ourmethod, \textit{w/o visual Memory}, \textit{w/o semantic Memory}, and \textit{w/o Dual-source Memory}. 
Table~\ref{tab:ablation_dual_memory} shows that the full dual-memory design achieves the best performance on both benchmarks, reaching \textbf{51.4}\% on MIKASA-Robo~\cite{cherepanov2026memory} and \textbf{97.0}\% on LIBERO-90~\cite{liu2023libero}.
Removing both memory streams causes the largest degradation, reducing success rates to 29.2\% and 92.1\%, respectively.
The degradation caused by removing either stream further verifies their complementary contributions.

\noindent\textbf{Latent-native Memory \textit{vs.} Policy-side Condition.}
We isolate the effect of memory consumption path in Table~\ref{tab:ablation_latent_native_memory}, where \textsc{Baseline} denotes the memory-free action policy without the visual and
semantic memory.
The baseline reaches only 29.2\% on MIKASA-Robo~\cite{cherepanov2026memory} and 92.1\% on Long-90~\cite{liu2023libero}, while adding memory as an external policy-side condition~\cite{shi2025memoryvla} improves the baseline performance.
Directly feeding raw retrieved evidence to the action policy also improves over the baseline, but remains below the full model.
Our \ourmethod achieves the highest success rates of \textbf{51.4}\% and \textbf{97.0}\%, outperforming policy-side conditioning by \textbf{10.2} and \textbf{2.2} points, respectively.
These results confirm that the gain comes from latent-native memory integration rather than merely adding memory as external context.


\noindent\textbf{The Number of Retrieved Memory Units $K$.}
We further study how the retrieval budget of the memory seeker affects manipulation performance by varying the number of retrieved memory units per vault (Eq.~\ref{eq:short_memory_retrieval}). 
As shown in Fig.~\ref{tab:ablation_retrieved_k}, performance improves as $K$ increases from 2 to 8, reaching \textbf{73.9}\% on SimplerEnv~\cite{li2025evaluating} and \textbf{97.0}\% on Long-90~\cite{liu2023libero}, but declines slightly at $K=12$.
This indicates that a moderate retrieval budget balances historical coverage against redundant evidence; therefore, we set $K=8$ by default.

\noindent\textbf{The Number of Latent Memory Tokens $L_\text{s}$ and $L_\text{l}$.}
We study the effects of visual- and semantic memory capacity in Fig.~\ref{fig:latentnum}. Increasing $L_\text{s}$ from 2 to 16 improves SimplerEnv~\cite{li2025evaluating} success rate from 61.4\% to 65.6\%, while the gain saturates at 32. With $L_\text{s}=8$, increasing $L_\text{l}$ further improves performance, reaching 73.9\% on SimplerEnv and 97.0\% on Long-90~\cite{liu2023libero}. 
Although larger latent memory token budgets can yield strong performance, they also lengthen the VLA self-attention context and increase computational costs; thus, we adopt $(L_\text{s},L_\text{l})=(8,4)$ as the default.






\section{Conclusion}
\label{sec:conclusion}
In this paper, we propose \ourmethod, a dual latent memory framework for reducing the temporal short-horizon bias of vision-language-action models. The core idea is to make robotic history part of the model's native latent context, rather than an external condition attached after multimodal reasoning.
\ourmethod organizes historical experience into complementary visual and semantic memory, and reconstructs retrieved evidence as compact latent tokens that directly participate in VLA reasoning and action generation.
Extensive evaluations across three simulation benchmarks and real-world manipulation tasks demonstrate its effectiveness, highlighting context-native latent memory as a promising direction for temporally aware VLA systems.

\input{}
{
    \small
    \bibliographystyle{ieeenat_fullname}
    \bibliography{main}

@String(CVPR= {IEEE Conf. Comput. Vis. Pattern Recog.})

@String(ICCV= {Int. Conf. Comput. Vis.})

@String(ICLR = {Int. Conf. Learn. Represent.})

@String(AAAI = {AAAI})

@String(CVPR  = {CVPR})

@String(ICCV  = {ICCV})

@String(ICLR  = {ICLR})

@article{black2024pi_0,
  title={{$\pi_0$}: A Vision-Language-Action Flow Model for General Robot Control},
  author={Black, Kevin and Brown, Noah and Driess, Danny and Esmail, Adnan and Equi, Michael and Finn, Chelsea and Fusai, Niccolo and Groom, Lachy and Hausman, Karol and Ichter, Brian and others},
  journal={arXiv preprint arXiv:2410.24164},
  year={2024}
}

@inproceedings{kim2025openvla,
  title={OpenVLA: An Open-Source Vision-Language-Action Model},
  author={Kim, Moo Jin and Pertsch, Karl and Karamcheti, Siddharth and Xiao, Ted and Balakrishna, Ashwin and Nair, Suraj and Rafailov, Rafael and Foster, Ethan P and Sanketi, Pannag R and Vuong, Quan and others},
  booktitle={CoRL},
  pages={2679--2713},
  year={2025}
}

@inproceedings{liu2025rdt,
  title={Rdt-1b: a diffusion foundation model for bimanual manipulation},
  author={Liu, Songming and Wu, Lingxuan and Li, Bangguo and Tan, Hengkai and Chen, Huayu and Wang, Zhengyi and Xu, Ke and Su, Hang and Zhu, Jun},
  booktitle={ICLR},
  volume={2025},
  pages={29982--30009},
  year={2025}
}

@inproceedings{zhen20243d,
  title={3D-VLA: A 3D Vision-Language-Action Generative World Model},
  author={Zhen, Haoyu and Qiu, Xiaowen and Chen, Peihao and Yang, Jincheng and Yan, Xin and Du, Yilun and Hong, Yining and Gan, Chuang},
  booktitle={ICML},
  pages={61229--61245},
  year={2024}
}

@article{cheang2024gr,
  title={Gr-2: A generative video-language-action model with web-scale knowledge for robot manipulation},
  author={Cheang, Chi-Lam and Chen, Guangzeng and Jing, Ya and Kong, Tao and Li, Hang and Li, Yifeng and Liu, Yuxiao and Wu, Hongtao and Xu, Jiafeng and Yang, Yichu and others},
  journal={arXiv preprint arXiv:2410.06158},
  year={2024}
}

@article{bai2023qwen,
  title={Qwen technical report},
  author={Bai, Jinze and Bai, Shuai and Chu, Yunfei and Cui, Zeyu and Dang, Kai and Deng, Xiaodong and Fan, Yang and Ge, Wenbin and Han, Yu and Huang, Fei and others},
  journal={arXiv preprint arXiv:2309.16609},
  year={2023}
}

@inproceedings{chen2024scaling,
  title={On scaling up a multilingual vision and language model},
  author={Chen, Xi and Djolonga, Josip and Padlewski, Piotr and Mustafa, Basil and Changpinyo, Soravit and Wu, Jialin and Ruiz, Carlos Riquelme and Goodman, Sebastian and Wang, Xiao and Tay, Yi and others},
  booktitle={CVPR},
  pages={14432--14444},
  year={2024}
}

@article{khazatsky2024droid,
  title={Droid: A large-scale in-the-wild robot manipulation dataset},
  author={Khazatsky, Alexander and Pertsch, Karl and Nair, Suraj and Balakrishna, Ashwin and Dasari, Sudeep and Karamcheti, Siddharth and Nasiriany, Soroush and Srirama, Mohan Kumar and Chen, Lawrence Yunliang and Ellis, Kirsty and others},
  journal={arXiv preprint arXiv:2403.12945},
  year={2024}
}

@article{chi2025diffusion,
  title={Diffusion policy: Visuomotor policy learning via action diffusion},
  author={Chi, Cheng and Xu, Zhenjia and Feng, Siyuan and Cousineau, Eric and Du, Yilun and Burchfiel, Benjamin and Tedrake, Russ and Song, Shuran},
  journal={The International Journal of Robotics Research},
  volume={44},
  number={10-11},
  pages={1684--1704},
  year={2025}
}

@inproceedings{zhang2025flowpolicy,
  title={Flowpolicy: Enabling fast and robust 3d flow-based policy via consistency flow matching for robot manipulation},
  author={Zhang, Qinglun and Liu, Zhen and Fan, Haoqiang and Liu, Guanghui and Zeng, Bing and Liu, Shuaicheng},
  booktitle={AAAI},
  volume={39},
  number={14},
  pages={14754--14762},
  year={2025}
}

@inproceedings{liu2023libero,
  title={Libero: Benchmarking knowledge transfer for lifelong robot learning},
  author={Liu, Bo and Zhu, Yifeng and Gao, Chongkai and Feng, Yihao and Liu, Qiang and Zhu, Yuke and Stone, Peter},
  booktitle={NeurIPS},
  volume={36},
  pages={44776--44791},
  year={2023}
}

@article{kim2025fine,
  title={Fine-tuning vision-language-action models: Optimizing speed and success},
  author={Kim, Moo Jin and Finn, Chelsea and Liang, Percy},
  journal={arXiv preprint arXiv:2502.19645},
  year={2025}
}

@inproceedings{songdenoising,
  title={Denoising Diffusion Implicit Models},
  author={Song, Jiaming and Meng, Chenlin and Ermon, Stefano},
  booktitle={International Conference on Learning Representations},
  year={2021}
}

@inproceedings{walke2023bridgedata,
  title={Bridgedata v2: A dataset for robot learning at scale},
  author={Walke, Homer Rich and Black, Kevin and Zhao, Tony Z and Vuong, Quan and Zheng, Chongyi and Hansen-Estruch, Philippe and He, Andre Wang and Myers, Vivek and Kim, Moo Jin and Du, Max and others},
  booktitle={CoRL},
  pages={1723--1736},
  year={2023}
}

@inproceedings{shi2025memoryvla,
  title={Memoryvla: Perceptual-cognitive memory in vision-language-action models for robotic manipulation},
  author={Shi, Hao and Xie, Bin and Liu, Yingfei and Sun, Lin and Liu, Fengrong and Wang, Tiancai and Zhou, Erjin and Fan, Haoqiang and Zhang, Xiangyu and Huang, Gao},
  booktitle={ICLR},
  year={2026}
}

@inproceedings{li2025evaluating,
  title={Evaluating Real-World Robot Manipulation Policies in Simulation},
  author={Li, Xuanlin and Hsu, Kyle and Gu, Jiayuan and Mees, Oier and Pertsch, Karl and Walke, Homer Rich and Fu, Chuyuan and Lunawat, Ishikaa and Sieh, Isabel and Kirmani, Sean and others},
  booktitle={CoRL},
  pages={3705--3728},
  year={2025},
  organization={PMLR}
}

@inproceedings{o2024open,
  title={Open x-embodiment: Robotic learning datasets and rt-x models: Open x-embodiment collaboration 0},
  author={O’Neill, Abby and Rehman, Abdul and Maddukuri, Abhiram and Gupta, Abhishek and Padalkar, Abhishek and Lee, Abraham and Pooley, Acorn and Gupta, Agrim and Mandlekar, Ajay and Jain, Ajinkya and others},
  booktitle={ICRA},
  pages={6892--6903},
  year={2024}
}

@article{li2024cogact,
  title={Cogact: A foundational vision-language-action model for synergizing cognition and action in robotic manipulation},
  author={Li, Qixiu and Liang, Yaobo and Wang, Zeyu and Luo, Lin and Chen, Xi and Liao, Mozheng and Wei, Fangyun and Deng, Yu and Xu, Sicheng and Zhang, Yizhong and others},
  journal={arXiv preprint arXiv:2411.19650},
  year={2024}
}

@inproceedings{huang2026thinkact,
  title={Thinkact: Vision-language-action reasoning via reinforced visual latent planning},
  author={Huang, Chi-Pin and Wu, Yueh-Hua and Chen, Min-Hung and Wang, Frank and Yang, Fu-En},
  booktitle={NeurIPS},
  pages={82782--82802},
  year={2025}
}

@article{bai2026latent,
  title={Latent Reasoning VLA: Latent Thinking and Prediction for Vision-Language-Action Models},
  author={Bai, Shuanghao and Lyu, Jing and Zhou, Wanqi and Li, Zhe and Wang, Dakai and Xing, Lei and Zhao, Xiaoguang and Wang, Pengwei and Wang, Zhongyuan and Chi, Cheng and others},
  journal={arXiv preprint arXiv:2602.01166},
  year={2026}
}

@inproceedings{octo_2023,
    title={Octo: An Open-Source Generalist Robot Policy},
    author = {{Octo Model Team} and Dibya Ghosh and Homer Walke and Karl Pertsch and Kevin Black and Oier Mees and Sudeep Dasari and Joey Hejna and Charles Xu and Jianlan Luo and Tobias Kreiman and {You Liang} Tan and Pannag Sanketi and Quan Vuong and Ted Xiao and Dorsa Sadigh and Chelsea Finn and Sergey Levine},
    booktitle = {RSS},
    year = {2024},
}

@article{qu2025spatialvla,
  title={Spatialvla: Exploring spatial representations for visual-language-action model},
  author={Qu, Delin and Song, Haoming and Chen, Qizhi and Yao, Yuanqi and Ye, Xinyi and Ding, Yan and Wang, Zhigang and Gu, JiaYuan and Zhao, Bin and Wang, Dong and others},
  journal={arXiv preprint arXiv:2501.15830},
  year={2025}
}

@article{pertsch2025fast,
  title={Fast: Efficient action tokenization for vision-language-action models},
  author={Pertsch, Karl and Stachowicz, Kyle and Ichter, Brian and Driess, Danny and Nair, Suraj and Vuong, Quan and Mees, Oier and Finn, Chelsea and Levine, Sergey},
  journal={arXiv preprint arXiv:2501.09747},
  year={2025}
}

@inproceedings{zhao2025cot,
  title={Cot-vla: Visual chain-of-thought reasoning for vision-language-action models},
  author={Zhao, Qingqing and Lu, Yao and Kim, Moo Jin and Fu, Zipeng and Zhang, Zhuoyang and Wu, Yecheng and Li, Zhaoshuo and Ma, Qianli and Han, Song and Finn, Chelsea and others},
  booktitle={CVPR},
  pages={1702--1713},
  year={2025}
}

@article{huang2026fast,
  title={Fast-ThinkAct: Efficient Vision-Language-Action Reasoning via Verbalizable Latent Planning},
  author={Huang, Chi-Pin and Man, Yunze and Yu, Zhiding and Chen, Min-Hung and Kautz, Jan and Wang, Yu-Chiang Frank and Yang, Fu-En},
  journal={arXiv e-prints},
  pages={arXiv--2601},
  year={2026}
}

@inproceedings{duan2025manipulate,
  title={Manipulate-Anything: Automating Real-World Robots using Vision-Language Models},
  author={Duan, Jiafei and Yuan, Wentao and Pumacay, Wilbert and Wang, Yi Ru and Ehsani, Kiana and Fox, Dieter and Krishna, Ranjay},
  booktitle={CoRL},
  pages={5326--5350},
  year={2025}
}

@inproceedings{huang2025rekep,
  title={ReKep: Spatio-Temporal Reasoning of Relational Keypoint Constraints for Robotic Manipulation},
  author={Huang, Wenlong and Wang, Chen and Li, Yunzhu and Zhang, Ruohan and Fei-Fei, Li},
  booktitle={CoRL},
  pages={4573--4602},
  year={2025}
}

@article{liu2025hybridvla,
  title={Hybridvla: Collaborative diffusion and autoregression in a unified vision-language-action model},
  author={Liu, Jiaming and Chen, Hao and An, Pengju and Liu, Zhuoyang and Zhang, Renrui and Gu, Chenyang and Li, Xiaoqi and Guo, Ziyu and Chen, Sixiang and Liu, Mengzhen and others},
  journal={arXiv preprint arXiv:2503.10631},
  year={2025}
}

@inproceedings{wang2025spec,
  title={Spec-vla: speculative decoding for vision-language-action models with relaxed acceptance},
  author={Wang, Songsheng and Yu, Rucheng and Yuan, Zhihang and Yu, Chao and Gao, Feng and Wang, Yu and Wong, Derek F},
  booktitle={EMNLP},
  pages={26916--26928},
  year={2025}
}

@inproceedings{zhang2026mole,
  title={Mole-vla: Dynamic layer-skipping vision language action model via mixture-of-layers for efficient robot manipulation},
  author={Zhang, Rongyu and Dong, Menghang and Zhang, Yuan and Heng, Liang and Chi, Xiaowei and Dai, Gaole and Du, Li and Wang, Dan and Du, Yuan and Zhang, Shanghang},
  booktitle={AAAI},
  volume={40},
  number={22},
  pages={18764--18772},
  year={2026}
}

@article{wang2025bitvla,
  title={Bitvla: 1-bit vision-language-action models for robotics manipulation},
  author={Wang, Hongyu and Xiong, Chuyan and Wang, Ruiping and Chen, Xilin},
  journal={arXiv preprint arXiv:2506.07530},
  year={2025}
}

@inproceedings{zitkovich2023rt,
  title={Rt-2: Vision-language-action models transfer web knowledge to robotic control},
  author={Zitkovich, Brianna and Yu, Tianhe and Xu, Sichun and Xu, Peng and Xiao, Ted and Xia, Fei and Wu, Jialin and Wohlhart, Paul and Welker, Stefan and Wahid, Ayzaan and others},
  booktitle={CoRL},
  pages={2165--2183},
  year={2023}
}

@article{cen2025worldvla,
  title={Worldvla: Towards autoregressive action world model},
  author={Cen, Jun and Yu, Chaohui and Yuan, Hangjie and Jiang, Yuming and Huang, Siteng and Guo, Jiayan and Li, Xin and Song, Yibing and Luo, Hao and Wang, Fan and others},
  journal={arXiv preprint arXiv:2506.21539},
  year={2025}
}

@article{wang2025unified,
  title={Unified vision-language-action model},
  author={Wang, Yuqi and Li, Xinghang and Wang, Wenxuan and Zhang, Junbo and Li, Yingyan and Chen, Yuntao and Wang, Xinlong and Zhang, Zhaoxiang},
  journal={arXiv preprint arXiv:2506.19850},
  year={2025}
}

@article{lin2025vote,
  title={Vote: vision-language-action optimization with trajectory ensemble voting},
  author={Lin, Juyi and Taherin, Amir and Akbari, Arash and Akbari, Arman and Lu, Lei and Chen, Guangyu and Padir, Taskin and Yang, Xiaomeng and Chen, Weiwei and Li, Yiqian and others},
  journal={arXiv preprint arXiv:2507.05116},
  year={2025}
}

@article{deng2025graspvla,
  title={Graspvla: a grasping foundation model pre-trained on billion-scale synthetic action data},
  author={Deng, Shengliang and Yan, Mi and Wei, Songlin and Ma, Haixin and Yang, Yuxin and Chen, Jiayi and Zhang, Zhiqi and Yang, Taoyu and Zhang, Xuheng and Zhang, Wenhao and others},
  journal={arXiv preprint arXiv:2505.03233},
  year={2025}
}

@inproceedings{wen2025dexvla,
  title={DexVLA: Vision-Language Model with Plug-In Diffusion Expert for General Robot Control},
  author={Wen, Junjie and Zhu, Yichen and Li, Jinming and Tang, Zhibin and Shen, Chaomin and Feng, Feifei},
  booktitle={CoRL},
  pages={3094--3114},
  year={2025}
}

@inproceedings{wang2025vq,
  title={Vq-vla: Improving vision-language-action models via scaling vector-quantized action tokenizers},
  author={Wang, Yating and Zhu, Haoyi and Liu, Mingyu and Yang, Jiange and Fang, Hao-Shu and He, Tong},
  booktitle={ICCV},
  pages={11089--11099},
  year={2025}
}

@article{sridhar2025memer,
  title={Memer: Scaling up memory for robot control via experience retrieval},
  author={Sridhar, Ajay and Pan, Jennifer and Sharma, Satvik and Finn, Chelsea},
  journal={arXiv preprint arXiv:2510.20328},
  year={2025}
}

@article{li2024towards,
  title={Towards Generalist Robot Policies: What Matters in Building Vision-Language-Action Models},
  author={Li, Xinghang and Li, Peiyan and Liu, Minghuan and Wang, Dong and Liu, Jirong and Kang, Bingyi and Ma, Xiao and Kong, Tao and Zhang, Hanbo and Liu, Huaping},
  journal={arXiv preprint arXiv:2412.14058},
  year={2024}
}

@inproceedings{fan2025interleave,
  title={Interleave-vla: Enhancing robot manipulation with interleaved image-text instructions},
  author={Fan, Cunxin and Jia, Xiaosong and Sun, Yihang and Wang, Yixiao and Wei, Jianglan and Gong, Ziyang and Zhao, Xiangyu and Tomizuka, Masayoshi and Yang, Xue and Yan, Junchi and others},
  booktitle={CoRL},
  year={2025}
}

@article{koo2025hamlet,
  title={Hamlet: Switch your vision-language-action model into a history-aware policy},
  author={Koo, Myungkyu and Choi, Daewon and Kim, Taeyoung and Lee, Kyungmin and Kim, Changyeon and Seo, Younggyo and Shin, Jinwoo},
  journal={arXiv preprint arXiv:2510.00695},
  year={2025}
}

@article{li2025cronusvla,
  title={CronusVLA: Towards Efficient and Robust Manipulation via Multi-Frame Vision-Language-Action Modeling},
  author={Li, Hao and Yang, Shuai and Chen, Yilun and Chen, Xinyi and Yang, Xiaoda and Tian, Yang and Wang, Hanqing and Wang, Tai and Lin, Dahua and Zhao, Feng and others},
  journal={arXiv preprint arXiv:2506.19816},
  year={2025}
}

@article{hu2026resolving,
  title={Resolving state ambiguity in robot manipulation via adaptive working memory recoding},
  author={Hu, Qingda and Qiu, Ziheng and Xu, Zijun and Zhang, Kaizhao and Bu, Xizhou and Sun, Zuolei and Zhang, Bo and Zhao, Jieru and Gan, Zhongxue and Ding, Wenchao},
  journal={IEEE Robotics and Automation Letters},
  year={2026}
}

@inproceedings{zheng2025tracevla,
  title={Tracevla: Visual trace prompting enhances spatial-temporal awareness for generalist robotic policies},
  author={Zheng, Ruijie and Liang, Yongyuan and Huang, Shuaiyi and Gao, Jianfeng and Daum{\'e} III, Hal and Kolobov, Andrey and Huang, Furong and Yang, Jianwei},
  booktitle={ICLR},
  volume={2025},
  pages={54277--54296},
  year={2025}
}

@article{li2026remem,
  title={ReMem-VLA: Empowering Vision-Language-Action Model with Memory via Dual-Level Recurrent Queries},
  author={Li, Hang and Shen, Fengyi and Chen, Dong and Yang, Liudi and Wang, Xudong and Shi, Jinkui and Bing, Zhenshan and Liu, Ziyuan and Knoll, Alois},
  journal={arXiv preprint arXiv:2603.12942},
  year={2026}
}

@article{mark2026bpp,
  title={Bpp: Long-context robot imitation learning by focusing on key history frames},
  author={Mark, Max Sobol and Liang, Jacky and Attarian, Maria and Fu, Chuyuan and Dwibedi, Debidatta and Shah, Dhruv and Kumar, Aviral},
  journal={arXiv preprint arXiv:2602.15010},
  year={2026}
}

@article{bu2025univla,
  title={Univla: Learning to act anywhere with task-centric latent actions},
  author={Bu, Qingwen and Yang, Yanting and Cai, Jisong and Gao, Shenyuan and Ren, Guanghui and Yao, Maoqing and Luo, Ping and Li, Hongyang},
  journal={arXiv preprint arXiv:2505.06111},
  year={2025}
}

@inproceedings{li2024vision,
  title={Vision-language foundation models as effective robot imitators},
  author={Li, Xinghang and Liu, Minghuan and Zhang, Hanbo and Yu, Cunjun and Xu, Jie and Wu, Hongtao and Cheang, Chilam and Jing, Ya and Zhang, Weinan and Liu, Huaping and others},
  booktitle={ICLR},
  volume={2024},
  pages={26703--26721},
  year={2024}
}

@inproceedings{peebles2023scalable,
  title={Scalable diffusion models with transformers},
  author={Peebles, William and Xie, Saining},
  booktitle={ICCV},
  pages={4195--4205},
  year={2023}
}

@inproceedings{karamcheti2024prismatic,
  title={Prismatic vlms: Investigating the design space of visually-conditioned language models},
  author={Karamcheti, Siddharth and Nair, Suraj and Balakrishna, Ashwin and Liang, Percy and Kollar, Thomas and Sadigh, Dorsa},
  booktitle={ICML},
  year={2024}
}

@article{oquabdinov2,
  title={DINOv2: Learning Robust Visual Features without Supervision},
  author={Oquab, Maxime and Darcet, Timoth{\'e}e and Moutakanni, Th{\'e}o and Vo, Huy V and Szafraniec, Marc and Khalidov, Vasil and Fernandez, Pierre and HAZIZA, Daniel and Massa, Francisco and El-Nouby, Alaaeldin and others},
  journal={Transactions on Machine Learning Research},
    year={2024}
}

@inproceedings{zhai2023sigmoid,
  title={Sigmoid loss for language image pre-training},
  author={Zhai, Xiaohua and Mustafa, Basil and Kolesnikov, Alexander and Beyer, Lucas},
  booktitle={ICCV},
  pages={11975--11986},
  year={2023}
}

@article{touvron2023llama,
  title={Llama 2: Open foundation and fine-tuned chat models},
  author={Touvron, Hugo and Martin, Louis and Stone, Kevin and Albert, Peter and Almahairi, Amjad and Babaei, Yasmine and Bashlykov, Nikolay and Batra, Soumya and Bhargava, Prajjwal and Bhosale, Shruti and others},
  journal={arXiv preprint arXiv:2307.09288},
  year={2023}
}

@inproceedings{hu2018squeeze,
  title={Squeeze-and-excitation networks},
  author={Hu, Jie and Shen, Li and Sun, Gang},
  booktitle={CVPR},
  pages={7132--7141},
  year={2018}
}

@article{bu2025agibot,
  title={Agibot world colosseo: A large-scale manipulation platform for scalable and intelligent embodied systems},
  author={Bu, Qingwen and Cai, Jisong and Chen, Li and Cui, Xiuqi and Ding, Yan and Feng, Siyuan and Gao, Shenyuan and He, Xindong and Hu, Xuan and Huang, Xu and others},
  journal={arXiv preprint arXiv:2503.06669},
  year={2025}
}

@article{li2024language,
  title={Language-guided object-centric diffusion policy for generalizable and collision-aware robotic manipulation},
  author={Li, Hang and Feng, Qian and Zheng, Zhi and Feng, Jianxiang and Chen, Zhaopeng and Knoll, Alois},
  journal={arXiv preprint arXiv:2407.00451},
  year={2024}
}

@article{ze20243d,
  title={3D Diffusion Policy: Generalizable Visuomotor Policy Learning via Simple 3D Representations},
  author={Ze, Yanjie and Zhang, Gu and Zhang, Kangning and Hu, Chenyuan and Wang, Muhan and Xu, Huazhe},
  journal={RSS},
  year={2024}
}

@inproceedings{li2026global,
  title={Global prior meets local consistency: Dual-memory augmented vision-language-action model for efficient robotic manipulation},
  author={Li, Zaijing and Hu, Bing and Shao, Rui and Chen, Gongwei and Jiang, Dongmei and Xie, Pengwei and Hao, Jianye and Nie, Liqiang},
  booktitle={CVPR},
  pages={35135--35145},
  year={2026}
}

@inproceedings{wang2026monet,
  title={Monet: Reasoning in Latent Visual Space Beyond Image and Language},
  author={Wang, Qixun and Shi, Yang and Wang, Yifei and Zhang, Yuanxing and Wan, Pengfei and Gai, Kun and Ying, Xianghua and Wang, Yisen},
  booktitle={CVPR},
  pages={12030--12040},
  year={2026}
}

@inproceedings{yang2026machine,
  title={Machine mental imagery: Empower multimodal reasoning with latent visual tokens},
  author={Yang, Zeyuan and Yu, Xueyang and Chen, Delin and Shen, Maohao and Gan, Chuang},
  booktitle={CVPR},
  pages={33510--33520},
  year={2026}
}

@inproceedings{li2025latent,
  title={Latent visual reasoning},
  author={Li, Bangzheng and Sun, Ximeng and Liu, Jiang and Wang, Ze and Wu, Jialian and Yu, Xiaodong and Chen, Hao and Barsoum, Emad and Chen, Muhao and Liu, Zicheng},
  booktitle={ICLR},
  year={2026}
}

@article{liu2026lara,
  title={LARA: Latent Action Representation Alignment for Vision-Language-Action Models},
  author={Liu, Mengya and Jia, Baoxiong and Huang, Jiangyong and Zhang, Jingze and Huang, Siyuan},
  journal={arXiv preprint arXiv:2606.07100},
  year={2026}
}

@inproceedings{ni2026semanticvla,
  title={SemanticVLA: Towards Semantic Reasoning over Action Memorization via Synergistic Explicit Trace and Latent Action Planning},
  author={Ni, Fei and Chen, Zhuo and Yuan, Yifu and Dong, Zibin and Yao, Xianze and Luo, Shan and Hao, Jianye and Deng, Jiankang and Zafeiriou, Stefanos},
  booktitle={CVPR},
  pages={12237--12247},
  year={2026}
}

@article{zhang2025memgen,
  title={Memgen: Weaving generative latent memory for self-evolving agents},
  author={Zhang, Guibin and Fu, Muxin and Yan, Shuicheng},
  journal={arXiv preprint arXiv:2509.24704},
  year={2025}
}

@inproceedings{li2026towards,
  title={Towards efficient and robust manipulation via multi-frame vision-language-action modeling},
  author={Li, Hao and Yang, Shuai and Chen, Yilun and Chen, Xinyi and Yang, Xiaoda and Tian, Yang and Wang, Hanqing and Wang, Tai and Lin, Dahua and Zhao, Feng and others},
  booktitle={AAAI},
  volume={40},
  number={22},
  pages={18388--18396},
  year={2026}
}

@inproceedings{zhang2026dreamvla,
  title={Dreamvla: a vision-language-action model dreamed with comprehensive world knowledge},
  author={Zhang, Wenyao and Liu, Hongsi and Qi, Zekun and Wang, Yunnan and Yu, Xinqiang and Zhang, Jiazhao and Dong, Runpei and He, Jiawei and Wang, He and Zhang, Zhizheng and others},
  booktitle={NeurIPS},
  pages={24195--24228},
  year={2025}
}

@inproceedings{yu2026vismem,
  title={Vismem: Latent vision memory unlocks potential of vision-language models},
  author={Yu, Xinlei and Xu, Chengming and Zhang, Guibin and Chen, Zhangquan and Zhang, Yudong and He, Yongbo and Jiang, Peng-Tao and Zhang, Jiangning and Hu, Xiaobin and Yan, Shuicheng},
  booktitle={CVPR},
  pages={31544--31555},
  year={2026}
}

@article{yu2026latent,
  title={The latent space: Foundation, evolution, mechanism, ability, and outlook},
  author={Yu, Xinlei and Chen, Zhangquan and He, Yongbo and Fu, Tianyu and Dong, Guanting and Yang, Cheng and Xu, Chengming and Ma, Yue and Hu, Xiaobin and Cao, Zhe and others},
  journal={arXiv preprint arXiv:2604.02029},
  year={2026}
}

@inproceedings{cherepanov2026memory,
  title={Memory, benchmark \& robots: A benchmark for solving complex tasks with reinforcement learning},
  author={Cherepanov, Egor and Kachaev, Nikita and Kovalev, Alexey and Panov, Aleksandr},
  booktitle={ICLR},
  volume={2026},
  pages={9759--9815},
  year={2026}
}

@inproceedings{liu2026ttf,
  title={Ttf-vla: Temporal token fusion via pixel-attention integration for vision-language-action models},
  author={Liu, Chenghao and Zhang, Jiachen and Li, Chengxuan and Zhou, Zhimu and Wu, Shixin and Huang, Songfang and Duan, Huiling},
  booktitle={AAAI},
  volume={40},
  number={22},
  pages={18452--18459},
  year={2026}
}
}


\end{document}


\maketitle

\appendix
\section*{Summary of the Appendix}
This supplementary material provides additional details and analyses for the
CVPR 2027 submission, titled \textit{``Dual-Source Latent Memory Weaving
for Vision-Language-Action Models''.}
The appendix is organized as follows:
\begin{itemize}
    \item \S\ref{supp:method_details} provides additional method details.
    \item \S\ref{supp:implementation} summarizes the architectural and training
    hyperparameters.
    \item \S\ref{supp:eva} provides the evaluation protocols for
    real-world and simulated environments.
    \item \S\ref{supp:ablation} reports additional quantitative results and ablation studies.
    \item \S\ref{supp:mech} provides the case study of the latent memory mechanism.
    \item \S\ref{supp:temporal_baselines} compares \ourmethod with
    representative temporal-context VLA baselines.
    \item \S\ref{supp:qualitative} presents additional qualitative results in
    real-world and simulated environments.
    \item \S\ref{sec:limitations} discusses the limitations and future work.
\end{itemize}

\section{Additional Method Details}
\label{supp:method_details}
\subsection{Dual Memory Vault Updating}
\label{app:vault_update}
The construction of the two memory vaults is described in the main paper.
Here, we provide the complete online updating procedure.
Both vaults are initialized as empty and reset at the beginning of each episode.
At each timestep $t$, \ourmethod first retrieves from the existing vaults and writes the current memory units only after predicting the corresponding action chunk.
The visual unit is the key--value pair $\bm{m}_\text{s}^t=(\bm{k}_\text{s}^t,\bm{v}_\text{s}^t)$ defined in Eq.~2, while the semantic unit is the current memory-grounded action representation $\bm{m}_\text{l}^t=\bm{H}^\text{action}$ produced by the action-reasoning pass.
Memory units are stored in chronological order, such that temporal adjacency
is determined directly by their positions in the vault. The current units are first appended to their corresponding vaults:
\begin{equation}
\label{eq:vault_append}
\begin{aligned}
\overline{\mathcal{M}}_t^\text{vis}
&=
\mathrm{Append}
\left(
    \mathcal{M}_{t-1}^\text{vis},
    \bm{m}_\text{s}^t
\right),
\\
\overline{\mathcal{M}}_t^\text{sem}
&=
\mathrm{Append}
\left(
    \mathcal{M}_{t-1}^\text{sem},
    \bm{m}_\text{l}^t
\right).
\end{aligned}
\end{equation}
Let
$n_\text{s}=\lvert\overline{\mathcal{M}}_t^\text{vis}\rvert$ and
$n_\text{l}=\lvert\overline{\mathcal{M}}_t^\text{sem}\rvert$.
Each vault has a maximum capacity of $L$ and is consolidated independently
only when its number of stored units exceeds this capacity.

Following the adjacent consolidation strategy of
MemoryVLA~\citep{shi2025memoryvla}, we identify the most similar pair of
neighboring units in each overflowing vault.
For the visual vault, similarity is computed between the stored retrieval
keys.
For the semantic vault, each action representation is mean-pooled to obtain
its comparison representation:
\begin{equation}
\label{eq:adjacent_memory_selection}
\begin{aligned}
i_\text{s}^*
&=
\underset{1\leq i<n_\text{s}}{\arg\max}\;
\mathrm{cos}
\left(
    \bm{k}_\text{s}^i,
    \bm{k}_\text{s}^{i+1}
\right),
\\
i_\text{l}^*
&=
\underset{1\leq i<n_\text{l}}{\arg\max}\,
\mathrm{cos}\!\left(
    \mathrm{MeanPool}(\bm{m}_\text{l}^i),
    \mathrm{MeanPool}(\bm{m}_\text{l}^{i+1})
\right).
\end{aligned}
\end{equation}
Restricting consolidation to adjacent units preserves the chronological
organization of each vault while removing locally redundant memory.

The selected units are merged using an unweighted arithmetic mean.
For the visual vault, the key and value are averaged separately, whereas
the semantic action representations are averaged directly:
\begin{equation}
\label{eq:dual_memory_merge}
\begin{aligned}
\widetilde{\bm{k}}_\text{s}
&=
\frac{1}{2}
\left(
    \bm{k}_\text{s}^{i_\text{s}^*}
    +
    \bm{k}_\text{s}^{i_\text{s}^*+1}
\right),
\\
\widetilde{\bm{v}}_\text{s}
&=
\frac{1}{2}
\left(
    \bm{v}_\text{s}^{i_\text{s}^*}
    +
    \bm{v}_\text{s}^{i_\text{s}^*+1}
\right),
\qquad
\widetilde{\bm{m}}_\text{s}
=
\left(
    \widetilde{\bm{k}}_\text{s},
    \widetilde{\bm{v}}_\text{s}
\right),
\\
\widetilde{\bm{m}}_\text{l}
&=
\frac{1}{2}
\left(
    \bm{m}_\text{l}^{i_\text{l}^*}
    +
    \bm{m}_\text{l}^{i_\text{l}^*+1}
\right).
\end{aligned}
\end{equation}
The retrieval representation of $\widetilde{\bm{m}}_\text{l}$ is subsequently
computed on demand using $\mathrm{MeanPool}(\widetilde{\bm{m}}_\text{l})$,
consistent with the semantic retrieval procedure in the main paper.

Let
$\mathrm{Replace}(\mathcal{M},i,\bm{m})$
denote replacing the $i$-th and $(i+1)$-th units in $\mathcal{M}$ with
$\bm{m}$ while leaving all remaining units in their original order.
The complete update is
\begin{equation}
\label{eq:complete_vault_update}
\begin{aligned}
\mathcal{M}_t^\text{vis}
&=
\begin{cases}
\overline{\mathcal{M}}_t^\text{vis},
& n_\text{s}\leq L,
\\
\mathrm{Replace}
\left(
    \overline{\mathcal{M}}_t^\text{vis},
    i_\text{s}^*,
    \widetilde{\bm{m}}_\text{s}
\right),
& n_\text{s}=L+1,
\end{cases}
\\[2mm]
\mathcal{M}_t^\text{sem}
&=
\begin{cases}
\overline{\mathcal{M}}_t^\text{sem},
& n_\text{l}\leq L,
\\
\mathrm{Replace}
\left(
    \overline{\mathcal{M}}_t^\text{sem},
    i_\text{l}^*,
    \widetilde{\bm{m}}_\text{l}
\right),
& n_\text{l}=L+1.
\end{cases}
\end{aligned}
\end{equation}
Since exactly one unit is appended to each vault at every policy step, its
size can exceed $L$ by at most one. A single consolidation therefore restores
the prescribed capacity.

\subsection{Latent Memory Seeker and Condenser}
\label{app:latent_memory_condenser}
\noindent\textbf{Timestep-aware memory retrieval.}
Following MemoryVLA~\citep{shi2025memoryvla}, each memory unit retains its
episode timestep $t_i$. Before retrieval, we add a fixed sinusoidal timestep
embedding $\mathrm{TE}(t_i)$ to the corresponding visual and semantic
retrieval keys:
\begin{equation}
\label{eq:timestep_aware_memory_keys}
\bar{\bm{k}}_\text{s}^{i}
=\bm{k}_\text{s}^{i}+\mathrm{TE}(t_i),
\qquad
\bar{\bm{k}}_\text{l}^{i}
=\mathrm{MeanPool}(\bm{m}_\text{l}^{i})+\mathrm{TE}(t_i).
\end{equation}
The augmented keys replace the original retrieval keys in the similarity
scores, while the stored value tokens remain unchanged. This encoding allows
the seeker to distinguish similar memory content occurring at different
stages of an episode.

\noindent\textbf{Query Builder $\mathcal{B}$.}
We implement the query builder $\mathcal{B}$ as a lightweight two-layer
Transformer that converts the current multimodal cognition representation
into a fixed-length retrieval query.
Given
$\bm{Q}_t^\text{con}\in\mathbb{R}^{N_\text{con}\times C}$,
we append the learnable query slots
$\bm{Q}^\text{init}\in\mathbb{R}^{K_\text{q}\times C}$
to form the input sequence:
\begin{equation}
\label{eq:query_builder_input}
\bm{U}_{\mathcal{B}}^{0}
=
[\bm{Q}_t^\text{con};\bm{Q}^\text{init}],
\qquad
\bm{Q}_t
=
\mathcal{B}
\left(
    \bm{U}_{\mathcal{B}}^{0};
    \bm{A}_{\mathcal{B}}
\right)[-K_\text{q}:].
\end{equation}
Each layer follows a standard pre-normalized Transformer block, consisting of
masked multi-head self-attention and a position-wise feed-forward network with
residual connections. The attention and feed-forward parameters are specific
to each module and Transformer layer.

The attention mask $\bm{A}_{\mathcal{B}}$ allows the appended query slots
to read the complete multimodal context while preventing the original
context tokens from attending to these slots. With rows indexing attention
queries and columns indexing attention keys, the mask is
\begin{equation}
\label{eq:query_builder_mask}
\left(\bm{A}_{\mathcal{B}}\right)_{ij}
=
\begin{cases}
-\infty, &
\begin{gathered}
1\leq i\leq N_\text{con},\\
N_\text{con}<j\leq N_\text{con}+K_\text{q},
\end{gathered}\\
0, & \text{otherwise}.
\end{cases}
\end{equation}
Thus, $\bm{Q}^\text{init}$ can attend to
$\bm{Q}_t^\text{con}$ and to the other query slots, whereas
$\bm{Q}_t^\text{con}$ can attend only within the original context
sequence. In implementation, $-\infty$ is replaced by a sufficiently
large negative value. After the final layer, the outputs corresponding to
the last $K_\text{q}$ positions are retained as $\bm{Q}_t$.

\noindent\textbf{Memory Formers $\mathcal{F}_\text{v}$ and $\mathcal{F}_\text{c}$.}
The visual memory former $\mathcal{F}_\text{v}$ and semantic memory former
$\mathcal{F}_\text{c}$ follow the same two-layer Transformer architecture
as $\mathcal{B}$. The two formers do not share parameters, and
$\mathcal{B}$ is instantiated separately. Moreover,
$\mathcal{F}_\text{v}$ and $\mathcal{F}_\text{c}$ use distinct learnable
target sequences
$\bm{T}_\text{s}\in\mathbb{R}^{L_\text{s}\times C}$ and
$\bm{T}_\text{l}\in\mathbb{R}^{L_\text{l}\times C}$, respectively.
Their inputs and outputs are
\begin{equation}
\label{eq:memory_former_details}
\begin{aligned}
\bm{U}_\text{v}^{0}
&= [\bm{Q}_t;\bm{Z}^\text{vis};\bm{T}_\text{s}],\\
\bm{M}^\text{vis}
&= \mathcal{F}_\text{v}
\left(\bm{U}_\text{v}^{0};\bm{A}_\text{v}\right)[-L_\text{s}:],\\
\bm{U}_\text{c}^{0}
&= [\bm{Q}_t;\bm{Z}^\text{sem};\bm{T}_\text{l}],\\
\bm{M}^\text{sem}
&= \mathcal{F}_\text{c}
\left(\bm{U}_\text{c}^{0};\bm{A}_\text{c}\right)[-L_\text{l}:].
\end{aligned}
\end{equation}

Let
$N_\text{v}=\lvert[\bm{Q}_t;\bm{Z}^\text{vis}]\rvert$ and
$N_\text{c}=\lvert[\bm{Q}_t;\bm{Z}^\text{sem}]\rvert$
denote the numbers of content tokens supplied to the two formers, and let
$L_\text{v}=L_\text{s}$ and $L_\text{c}=L_\text{l}$.
For $x\in\{\text{v},\text{c}\}$, the corresponding attention mask is
\begin{equation}
\label{eq:memory_former_mask}
\left(\bm{A}_x\right)_{ij}
=
\begin{cases}
-\infty, &
\begin{gathered}
1\leq i\leq N_x,\\
N_x<j\leq N_x+L_x,
\end{gathered}\\
0, & \text{otherwise}.
\end{cases}
\end{equation}
Under this mask, the context query $\bm{Q}_t$ and the retrieved memory
tokens can attend to one another. The target slots can attend to
$\bm{Q}_t$, all retrieved memory tokens, and the other target slots, while
neither $\bm{Q}_t$ nor the retrieved memory tokens can attend back to the
target slots. The outputs at the target positions therefore provide
fixed-length latent memory tokens conditioned jointly on the current
multimodal context and the retrieved historical evidence.

\section{Implementation and Training Details}
\label{supp:implementation}
\subsection{Hyper-parameters}
\label{app:hyperparameters}

We summarize the principal architectural and optimization hyper-parameters
of \ourmethod in Tables~\ref{tab:architecture_hyperparameters}
and~\ref{tab:training_hyperparameters}.
Unless otherwise specified, the same model and memory configurations are
used across all benchmarks.
We instantiate \ourmethod with a 7B-parameter Prismatic VLM and a diffusion
action expert containing approximately 300M parameters.
The policy receives a single third-person RGB observation resized to
$224\times224$ and predicts a chunk of 16 continuous 7-DoF actions.
Both memory vaults have a capacity of $L=16$, from which the memory seeker
retrieves $K=8$ units per vault.
The memory condenser produces $L_\text{s}=8$ visual and
$L_\text{l}=4$ semantic latent memory tokens.
The query builder $\mathcal{B}$ and memory formers
$\mathcal{F}_\text{v}$ and $\mathcal{F}_\text{c}$ each contain two
Transformer layers and are parameterized independently.
During inference, the diffusion action expert uses DDIM sampling with
10 denoising steps.

\begin{table}[t]
\centering
\caption{Architecture and memory hyper-parameters of \ourmethod
(\S\ref{app:hyperparameters}).}
\label{tab:architecture_hyperparameters}
\small
\setlength{\tabcolsep}{5pt}
\renewcommand{\arraystretch}{1.08}
\begin{tabular}{@{}>{\raggedright\arraybackslash}p{0.55\linewidth}
                   >{\raggedright\arraybackslash}p{0.39\linewidth}@{}}
\toprule
Hyper-parameter & Value \\
\midrule

\multicolumn{2}{@{}l}{\textit{Backbone and input}} \\[1mm]

Vision-language backbone
& Prismatic VLM \\

VLM parameter count
& 7B \\

Visual encoders
& DINOv2 and SigLIP \\

VLA hidden dimension $C$
& 4096 \\

Camera view
& Single third-person RGB \\

Model input resolution
& $224\times224$ \\

Raw real-world image resolution
& $640\times480$ \\

Visual-token length $N_\text{i}$
& 128 \\

Action dimension
& 7-DoF \\

Action horizon $H$
& 16 \\

Number of action queries $N_\text{a}$
& 4 \\

Diffusion action expert
& DiT-L \\

Diffusion expert parameter count
& Approximately 300M \\

\midrule
\multicolumn{2}{@{}l}{\textit{Memory curator and vaults}} \\[1mm]

visual compressor $\mathcal{C}_\text{s}$
& SE-bottleneck \\

SE reduction ratio
& 16 \\

Capacity per vault $L$
& 16 \\

Memory write frequency
& One unit per policy step \\

\midrule
\multicolumn{2}{@{}l}{\textit{Memory seeker and condenser}} \\[1mm]

Retrieval budget $K$
& 8  \\

Number of query slots $K_\text{q}$
& 8 \\

visual output length $L_\text{s}$
& 8 tokens \\

semantic output length $L_\text{l}$
& 4 tokens \\

Depth of $\mathcal{B}$
& 2 Transformer layers \\

Depth of $\mathcal{F}_\text{v}$ and $\mathcal{F}_\text{c}$
& 2 Transformer layers each \\

\bottomrule
\end{tabular}
\end{table}

\begin{table}[t]
\centering
\caption{Optimization and diffusion hyper-parameters of \ourmethod
(\S\ref{app:hyperparameters}).}
\label{tab:training_hyperparameters}
\small
\setlength{\tabcolsep}{5pt}
\renewcommand{\arraystretch}{1.08}
\begin{tabular}{@{}>{\raggedright\arraybackslash}p{0.55\linewidth}
                   >{\raggedright\arraybackslash}p{0.39\linewidth}@{}}
\toprule
Hyper-parameter & Value \\
\midrule

\multicolumn{2}{@{}l}{\textit{Optimization}} \\[1mm]

Training accelerators
& 8 NVIDIA H800 GPUs \\

Distributed training strategy
& PyTorch FSDP \\

Batch size per GPU
& 32 \\

Global batch size
& 256 \\

Optimizer
& AdamW \\

Learning rate
& $2\times10^{-5}$ \\

Maximum gradient norm
& 1.0 \\

Exponential moving average
& Disabled \\

\midrule
\multicolumn{2}{@{}l}{\textit{Diffusion action generation}} \\[1mm]

Training objective
& Mean squared error \\

Repeated diffusion samples per training input
& 4 \\

Training diffusion timesteps
& 100 \\

Noise schedule
& Squared cosine \\

Inference sampler
& DDIM \\

Inference denoising steps
& 10 \\

Classifier-free guidance scale
& 1.5 \\

DDIM stochasticity $\eta$
& 0.0 \\

Denoised-output clipping
& Disabled \\

\bottomrule
\end{tabular}
\end{table}
\subsection{Training Setup}
\label{app:memory_training_samples}

Offline demonstrations are organized into episode-consistent sequences rather
than independent frames. Within each sequence, frames are processed in temporal
order while the two vault states are carried forward; the vaults are reset at
episode or independently sampled group boundaries. Thus, each ordered sequence
and its evolving vault states form one stateful memory training sample.

\noindent\textbf{SimplerEnv-Bridge.}
We train on Bridge v2 for $50$K steps using a stream dataloader, following the
episode-aware design of MemoryVLA~\citep{shi2025memoryvla}. Each demonstration
is unpacked into consecutive frames tagged with its episode ID. A minibatch is
filled sequentially from the same episode whenever possible. If an episode
ends before the minibatch is full, the remaining positions are filled from the
next episode and the vaults are reset at the episode boundary. The next
minibatch resumes from the previous stream cursor, preserving the temporal
order within each episode. The memory length is set to $16$.

\noindent\textbf{LIBERO.}
Following OpenVLA~\citep{kim2025openvla}, we use $50$ demonstrations per task.
Spatial, Object, and Goal are trained separately for $20$K steps each, whereas
Long-10 and Long-90 are trained jointly for $40$K steps. Each minibatch is
divided into temporal groups, with each group containing $16$ frames sampled
from a single episode and sorted by their original timestamps. The vault states
are propagated within each group and reset between groups. The memory length is
set to $16$.

\noindent\textbf{MIKASA-Robo.}
We jointly train the five tasks on all $1{,}250$ demonstrations for $20$K steps,
using $128\times128$ RGB observations and delta end-effector control. We adopt
the same grouped dataloader as LIBERO and set the memory length to $16$.

\noindent\textbf{Real-world.}
Models are trained for $5$K--$20$K steps depending on the task. The General
Tasks and Long-horizon Tasks contain $200$ and $300$ demonstrations per task,
respectively. The memory length is set to 16 for general tasks and 64 for
long-horizon temporal tasks.

\subsection{Data Augmentation}
\label{app:data_augmentation}

During training, each third-person RGB observation is processed by a fixed sequence of stochastic augmentations: random resized cropping, followed by brightness, contrast, saturation, and hue jittering. The crop covers $90\%$ of the original image area with a fixed aspect ratio of $1.0$ and is subsequently resized to $224\times224$. We set the brightness jitter magnitude to $0.2$, sample the contrast and saturation factors uniformly from $[0.8,1.2]$, and sample the hue offset uniformly from $[-0.05,0.05]$. All augmentations are disabled during evaluation.

\begin{table*}[t]
\centering
\caption{Overview of \textit{General Tasks} in real-world evaluation (\S\ref{supp:real}).}
\label{tab:real_world_generalization_tasks}
\vspace{-4pt}
\small
\setlength{\tabcolsep}{7pt}
\renewcommand{\arraystretch}{1.15}

\begin{tabularx}{\textwidth}{
    >{\raggedright\arraybackslash}p{0.17\textwidth}
    *{3}{>{\raggedright\arraybackslash}X}
}
\toprule
& \multicolumn{3}{c}{\textbf{General Tasks}} \\
\cmidrule(lr){2-4}

\textbf{Task name}
& Stack cup
& Drawer
& Pick fruit \\
\midrule

\textbf{Prompt}
& Stack one cup on top of another cup.
& Open the drawer, put the items on the table inside, and close the drawer.
& Place the banana, kiwi, and wax apple into the basket in that order. \\

\textbf{\# demonstrations}
& 200 & 200 & 200 \\

\textbf{\# rollouts}
& 50 & 50 & 50 \\

\textbf{Performance (\%)}
& 82 & 78 & 72 \\
\bottomrule
\end{tabularx}
\end{table*}

\begin{table*}[t]
\centering
\caption{Overview of \textit{Long-horizon Tasks} in real-world evaluation (\S\ref{supp:real}).}
\label{tab:real_world_long_horizon_tasks}
\vspace{-4pt}
\small
\setlength{\tabcolsep}{7pt}
\renewcommand{\arraystretch}{1.15}

\begin{tabularx}{\textwidth}{
    >{\raggedright\arraybackslash}p{0.17\textwidth}
    *{3}{>{\raggedright\arraybackslash}X}
}
\toprule
& \multicolumn{3}{c}{\textbf{Long-horizon Tasks}} \\
\cmidrule(lr){2-4}

\textbf{Task name}
& Cover block
& Press buttons
& Fill drawer \\
\midrule

\textbf{Prompt}
& Cover the building block with a cup, then lift up the cup covering the block.
& Press and hold the red and blue buttons in sequence for a while.
& Put the three items on the table into the drawer, then close the drawer. \\

\textbf{\# demonstrations}
& 300 & 300 & 300 \\

\textbf{\# rollouts}
& 50 & 50 & 50 \\

\textbf{Performance (\%)}
& 56 & 66 & 62 \\
\bottomrule
\end{tabularx}
\end{table*}

\begin{table*}[t]
\centering
\caption{Task descriptions (\S\ref{supp:real}).}
\label{tab:real_world_task_descriptions}
\vspace{-4pt}
\small
\setlength{\tabcolsep}{7pt}
\renewcommand{\arraystretch}{1.15}

\begin{tabularx}{\textwidth}{
    >{\raggedright\arraybackslash}p{0.17\textwidth}
    >{\raggedright\arraybackslash}X
}
\toprule
\textbf{Task} & \textbf{Description} \\
\midrule

Stack cup
& Pick up one paper cup and stack it on top of the other cup. \\

Drawer
& Open the drawer, put the items on the table into it, and close the drawer. \\

Pick fruit
& Pick up the banana, the kiwi, and the wax apple and place them into the basket in this order. \\

Cover block
& Pick up a cup, place it upside down over the building block so that the block is hidden, and then lift the same cup to reveal the block. \\

Press buttons
& Press and hold the red button for a while, and then press and hold the blue button for a while. \\

Fill drawer
& Put the three items on the table into the open drawer one by one, and then close the drawer. \\

\bottomrule
\end{tabularx}
\end{table*}
\section{Additional Evaluation Details}
\label{supp:eva}

\subsection{Evaluation on SimplerEnv-Bridge}
We follow the official CogACT evaluation protocol~\citep{li2024cogact}, including its evaluation scripts and adaptive action ensemble strategy, with $\alpha=0.1$ and an ensemble horizon of $7$. We train each model for $50$K steps and evaluate every $2.5$K steps, selecting the checkpoint with the best validation performance since the diffusion training objective is not a reliable proxy for policy performance. The selected checkpoint is evaluated over 24 trials per task. We additionally re-evaluate the released CogACT-Large checkpoint under the same protocol, as its per-task results were not provided in the original paper. For $\pi_0$, we report the float32 results from the open-source reproduction with both uniform and Beta timestep sampling.

\subsection{Evaluation on MIKASA-Robo}
For MIKASA-Robo~\citep{cherepanov2026memory}, we evaluate each task over 100
episodes. Validation is performed every $1$K training steps, and we report the
success rate achieved by the checkpoint with the best validation performance.
As in LIBERO, we do not use action ensembling in our MIKASA-Robo experiments.

\subsection{Evaluation on LIBERO}
We evaluate on all five LIBERO suites~\citep{liu2023libero}, including Spatial, Object, Goal, Long-10, and Long-90. Models are validated every $1$K steps, and each task is evaluated over $50$ trials, with success rates reported at the best validation checkpoint. Unlike SimplerEnv-Bridge, we do not apply action ensembling on LIBERO. For a fair comparison, we reproduce CogACT~\citep{li2024cogact} using its official codebase, while results for $\pi_0$~\cite{black2024pi_0} and $\pi_0$-FAST~\citep{pertsch2025fast} are taken from their reported numbers. Notably, both $\pi_0$ variants additionally use wrist-camera observations and proprioceptive states, whereas our method relies solely on a single third-person RGB stream. Despite this more restricted observation setting, our method achieves higher success rates across the evaluated suites.

\SuppInput{tabs/inference}
\subsection{Real-world Evaluation}
\label{supp:real}

\noindent\textbf{Platform.} We use the AgileX Piper, a 6-DoF desktop manipulator mounted on a fixed table. Perception is provided by one exocentric RealSense D435i RGB camera placed in front of the workspace; depth is not used. Demonstrations are collected by human teleoperation at 30 Hz and stored as raw $640\times480$ frames with synchronized joint positions, end-effector poses, and gripper openings. During deployment, the policy receives the current RGB frame and the language instruction, and predicts an action chunk of 7-D end-effector deltas (translation, rotation, gripper), which are converted to joint targets by inverse kinematics and executed at 30 Hz.

\noindent\textbf{Task Suites.}
As summarized in Tables~\ref{tab:real_world_generalization_tasks} and~\ref{tab:real_world_long_horizon_tasks}, we construct two real-world task suites: General Tasks and Long-horizon Tasks. General Tasks evaluate standard pick-and-place and articulated-object manipulation under randomized object configurations, where task progress can be inferred directly from the current observation. In contrast, Long-horizon Tasks assess the ability to preserve and exploit historical context, as at least one decision cannot be resolved from the current observation alone due to object occlusion, visually aliased task states, or elapsed-time-dependent requirements. For each evaluation episode, object poses are randomly initialized within the workspace. The test-time language instruction is identical to that used during data collection. Detailed task descriptions and prompts are provided in Table~\ref{tab:real_world_task_descriptions}, with additional qualitative results presented in the next section.

\section{Additional Ablation Studies}
\label{supp:ablation}

\noindent\textbf{Inference Efficiency.}
We further compare the inference efficiency of \ourmethod with its CogACT baseline~\cite{li2024cogact} under the same inference configuration. As shown in Table~\ref{tab:inference_efficiency}, \ourmethod requires 0.315s to generate a 16-step action chunk, corresponding to an action-generation throughput of 50.8Hz. Compared with CogACT, the incorporation of latent memory increases latency by 0.076s and reduces throughput by 16.1Hz. Meanwhile, peak GPU memory increases from 15.9GB to 17.8GB, corresponding to an additional overhead of 1.9GB. The overall computational overhead remains controlled by the bounded design of our latent memory. Moreover, adjacent similarity merging prevents the vault size from growing with the trajectory length. These results show that \ourmethod introduces long-horizon memory reasoning into the VLA model while retaining an action throughput of 50.8Hz with controlled additional latency and GPU memory consumption.
\SuppInput{tabs/memory_size}
\SuppInput{tabs/memory_merge}

\noindent\textbf{Memory Bank Size $L$.}
We next investigate the impact of memory bank size $L$, which controls the maximum number of units maintained in each memory vault. As shown in Fig.~\ref{fig:size}, increasing $L$ from 4 to 16 consistently improves the success rate from 94.8\% to 97.0\% on Long-90~\cite{liu2023libero} and from 68.7\% to 73.9\% on SimplerEnv~\cite{li2025evaluating}. This supports our hypothesis that sufficient capacity is essential for preserving diverse historical evidence and task-progress cues, whereas an overly small memory bank compresses the history too aggressively. However, further increasing $L$ to 32 or 64 yields only a marginal gain of 0.2 points on Long-90 and instead decreases performance by 2.0 points on SimplerEnv. We speculate that an excessively large memory bank delays the consolidation of locally redundant states and introduces unnecessary candidates into memory retrieval, without providing additional decision-relevant evidence. Therefore, we set $L=16$ as the default configuration, which strikes a favorable balance between historical information preservation, retrieval effectiveness, and memory efficiency.
\SuppInput{tabs/temporal_comp}
\SuppInput{figs/attention_mech}

\noindent\textbf{Memory Vault Updating Strategy.}
We further investigate the impact of different memory vault updating strategies.
We compare our strategy with two alternatives: FIFO discards the earliest memory unit once the vault reaches its capacity, while Random Merge randomly selects memory units for consolidation without considering their representation similarity.
All variants use the same fixed capacity for both memory vaults.
As shown in Fig.~\ref{fig:strategy}, our adjacent similarity merging consistently achieves the best performance across all five LIBERO suites~\cite{liu2023libero}.
Notably, larger gains are observed on Goal, Long-10, and Long-90, where \ourmethod outperforms FIFO by \textbf{4.4}, \textbf{5.2}, and \textbf{3.4} points and Random Merge by \textbf{4.0}, \textbf{5.4}, and \textbf{3.0} points, respectively.
FIFO may prematurely remove early evidence that remains relevant to subsequent decisions, whereas Random Merge may combine distinct historical states and thereby obscure informative temporal cues.
In contrast, our strategy consolidates the most similar pair among temporally adjacent memory units, preferentially removing local redundancy while retaining distinctive state transitions.
These results verify that similarity-aware local consolidation more effectively preserves task-relevant historical information, particularly when action prediction depends on task progress over extended horizons.

\section{Case Study of Memory Mechanism}
\label{supp:mech}
Fig.~\ref{fig:attention} visualizes the retrieved memory elements and their attention weights in the real-world Cover Block task. Once the block is covered by a cup, \ourmethod assigns high attention to earlier frames revealing its original location, preserving the correspondence between the hidden block and the covering cup. This demonstrates that \ourmethod retrieves decision-relevant historical information to guide subsequent actions when critical visual information is absent from the current observation.
\SuppInput{figs/case_general}
\SuppInput{figs/case_long}
\SuppInput{figs/case_libero}

\section{Comparison with Temporal-Context VLA Baselines}
\label{supp:temporal_baselines}
We conduct a focused comparison with representative temporal-context VLAs that explicitly exploit historical observations or memory. As summarized in Table~\ref{tab:temporal_baselines}, \ourmethod consistently achieves the best average performance across SimplerEnv-Bridge~\citep{li2025evaluating}, LIBERO~\citep{liu2023libero}, and Mikasa-Robo~\citep{cherepanov2026memory}. On SimplerEnv-Bridge, it reaches 73.9\%, outperforming MemoryVLA~\citep{shi2025memoryvla} and CronusVLA~\citep{li2026towards} by \textbf{2.0} and \textbf{13.5} points, respectively. On LIBERO~\cite{liu2023libero}, \ourmethod achieves 97.7\% over the four commonly reported suites, exceeding the strongest temporal-context baseline by \textbf{0.7} points. More notably, on the memory-intensive Mikasa-Robo benchmark~\cite{cherepanov2026memory}, it improves over MemoryVLA and CronusVLA by \textbf{10.2} and \textbf{33.4} points, respectively. These consistent gains across diverse temporal dependencies substantiate the effectiveness of integrating historical evidence as context-native latent memory for action reasoning.

\section{Qualitative Results}
\label{supp:qualitative}
\subsection{Real-world Evaluation}
We present qualitative real-world results of \ourmethod. By weaving complementary visual and semantic memories into VLA reasoning, \ourmethod exhibits superior performance on both General Tasks (Fig.~\ref{fig:casegen}) and Long-horizon Tasks (Fig.~\ref{fig:caselong}).

\subsection{Simulation Evaluation}
Qualitative results in simulated environments are presented in Fig.~\ref{fig:caselibero}. Specifically, Fig.~\ref{fig:caselibero} provides representative trajectories on LIBERO~\cite{liu2023libero}, spanning all five suites.

\section{Limitations and Future Work}
\label{sec:limitations}

Despite its effectiveness, \ourmethod has two limitations that merit further investigation. \textbf{First}, the current memory curator operates with fixed storage and retrieval budgets, and consolidates adjacent memory units according to their representation similarity. This design keeps the memory overhead bounded, but implicitly treats similarity as a proxy for future utility. In long or event-dense trajectories, evidence that appears redundant when stored may later become essential for resolving task progress, while irreversible consolidation makes such information difficult to recover. A promising direction is to develop task-conditioned memory management that jointly determines what to retain, merge, and retrieve according to the estimated decision value and uncertainty of each memory unit. Dynamically allocating memory capacity based on trajectory complexity may further improve the balance between information preservation and efficiency. \textbf{Second}, the current memory is confined to a single episode and reset once the episode terminates. This prevents information leakage across trajectories, but also means that the policy cannot reuse experience accumulated from previous interactions. Future work may extend \ourmethod with persistent experiential memory, yielding a hierarchical memory system that couples within-episode working memory with reusable cross-episode experience. Such an extension requires reliable mechanisms for online consolidation, selective forgetting, and uncertainty-aware retrieval, so that relevant experience can facilitate new tasks without introducing negative transfer from outdated or mismatched memories.

\IfFileExists{main.bib}{%
  {\small
    \bibliographystyle{ieeenat_fullname}
    \bibliography{main}
  }%
}{%
  \PackageWarningNoLine{lamem-supp}{Missing main.bib; bibliography omitted}%
}